\documentclass[conference]{IEEEtran}
\IEEEoverridecommandlockouts
\usepackage{cite}
\usepackage{amsmath,amssymb,amsfonts}
\usepackage{algorithmic}
\usepackage{graphicx}
\usepackage{textcomp}
\usepackage{xcolor}
\usepackage{comment}
\usepackage{epsfig}
\usepackage{url}
\usepackage{enumitem}

\def\BibTeX{{\rm B\kern-.05em{\sc i\kern-.025em b}\kern-.08em
    T\kern-.1667em\lower.7ex\hbox{E}\kern-.125emX}}
\begin{document}

\title{\textit{OWL}: A Novel Approach to Machine Perception During Motion}

\author{\IEEEauthorblockN{Daniel Raviv}
\IEEEauthorblockA{\textit{Electrical Engineering and Computer Science Department} \\
\textit{Florida Atlantic University}\\
Boca Raton, Florida, USA \\
email: ravivd@fau.edu}
\and
\IEEEauthorblockN{Juan D. Yepes}
\IEEEauthorblockA{\textit{Electrical Engineering and Computer Science Department} \\
\textit{Florida Atlantic University}\\
Boca Raton, Florida, USA \\
email: jyepes@fau.edu}
}

\maketitle


\begin{abstract}
We introduce a perception-related function, \textit{OWL}, designed to address the complex challenges of 3D perception during motion. It derives its values directly from two perceived, fundamental visual motion cues, with one set of cue values per point per time instant. During motion, two perceived visual motion cues relative to a fixation point emerge: 1) perceived local visual looming of points near the fixation point, and 2) perceived rotation of the rigid object relative to the fixation point. It also expresses the relation between two well-known physical quantities, namely the relative instantaneous directional range and the directional translation expressions in 3D between the camera and any visible 3D point. This is accomplished \textit{without} explicitly requiring their measurement, any prior knowledge or calculation of their individual values. The function offers a unified, overarching, and analytical time-based approach that enhances and simplifies key perception capabilities, including scaled 3D mapping and camera heading. Simulation results demonstrate that the \textit{OWL} function achieves geometric constancy of 3D objects over time and enables scaled 3D scene reconstruction from visual motion cues alone. By leveraging simple and direct measurements from raw visual motion image sequences, the values of the new \textit{OWL} function can be obtained without requiring prior knowledge of stationary environments, moving objects, or the camera motion. 

This approach employs minimalistic, pixel-based, parallel computations, providing an alternative real-time representation for 3D points in relative motion. The \textit{OWL} function bridges the gap between theoretical concepts and practical applications in robotics and autonomous navigation. By advancing the boundaries of 3D perception during motion, the \textit{OWL} function may unlock new possibilities for applications requiring real-time decision-making and interaction, potentially establishing itself as a fundamental building block for next-generation autonomous systems. This paper offers an alternative perspective on machine perception, with implications that may extend to aspects of natural perception that could contribute to a better understanding of behavioral psychology and neural functionality.
\end{abstract}

\section{Introduction}
\subsection{Observations}
Have you ever wondered how a fly, with its tiny brain, manages to navigate and avoid collisions - even when surrounded by other moving objects or creatures, whether friend or foe? While we don’t yet have a definitive answer, this question has inspired our research: can we create simple, perhaps even basic, sensing-based representations and strategies that allow machines to "think like a fly"?. 
In flies, this relationship is astonishingly clear and rapid, pointing to an inherent simplicity in how they sense and respond. Flies transform low-resolution sequences of images into actionable signals, demonstrating a direct and highly correlated realtime connection between sensing, processing, and action.

Now, consider a gamer interacting with a 2D, continuously changing projection of a simulated 3D artificial world. They skillfully navigate this environment by swiftly responding to changes in the 2D projection, which, among other things, simulate perceptual threats from both stationary and moving elements within the artificially rendered 3D space. From this gaming scenario, several insights emerge: 

\begin{enumerate}[label={(\arabic*)}, leftmargin=*]
\item Variations in screen size, viewing angle, or position relative to the screen do not significantly influence the gameplay experience. 
\item The information required to play is derived entirely from a 2D sequence, without relying on actual 3D depth information or stereoscopic vision. This suggests that acting on 3D information is unnecessary for certain perception-based tasks and that 2D data alone is sufficient for some navigation and control tasks.
\item The gamer can begin playing without any prior knowledge of the environment, whether it involves flying through a canyon or racing a simulated car.
\item The cognitive process in the gamer’s mind appears computationally simple and robust. Despite the continuously changing 2D projections, the gamer perceives the world and moving objects as 3D rigid bodies, i.e., they perceived 3D constancy. Remarkably, gamers can play for hours without experiencing fatigue, suggesting minimal perceptual demands.
\end{enumerate}
This raises a fundamental question: \textbf{Are there perceived visual motion cues that are scale- and depth-independent and can be directly derived from continuously changing 2D visual information? Can they serve as a basis for a new perception domain?}

\subsection{About the Approach}
Two visual motion cues can be related to our own experiences: Imagine you're driving in a city and observing another vehicle moving relative to you. With one eye, fixate on a single point on that vehicle. Its optical flow is zero. However, two critical visual cues can still be derived from the changing projections of the surrounding points.

\begin{enumerate}
    \item \textbf{Perceived visual looming} -- due to changes in relative range between you and the fixation point on the vehicle.
    \item \textbf{Perceived visual rotation} -- due to the relative motion of the 3D vehicle relative to the camera (observer).
\end{enumerate}

These two cues form the foundation, i.e., the fundamental building block, of our approach. They are instantaneous, per point, visual motion indicators that enable perceived constancy of the 3D stationary environment and dynamic objects.

In addition, they can be derived independently of the 3D point for which they are desired, obtained entirely from points in a 2D sequence, and processed in parallel for all points regardless of the field of view or the 3D environment.

The values of these two perceived cues can be combined to obtain a 2D complex function that we call \textit{OWL}. \textit{OWL} can also be referred to as a 2D complex domain that uses the above two visual cues. The new approach leverages raw visual data and provides a unified, analytical and simplified, time-based framework, thus enhancing real-time perception and action capabilities, such as 3D mapping, i.e., shape constancy under relative motion (up to a scale factor). In addition, it can be used for finding relative heading and obstacle avoidance. The values of the \textit{OWL} function over time, as represented in the \textit{OWL} domain, can be used to segment moving objects, predict future location, define safe space, and allow for decisions about actions. However, these are beyond the scope of this paper. In this paper we focus on 3D perception-based reconstruction and heading.

Using \textit{OWL}, stationary 3D objects appear geometrically unchanged, thereby maintaining scaled shape constancy over time despite the large volume of dynamic visual data. This facilitates a simple and efficient 3D scaled reconstruction of the scene when relative motion is present.

\textit{OWL} values can be represented in the complex domain for the 2D case and extended using quaternions for the 3D case. In this paper, for basic 2D explanations, we start with complex numbers, and for comprehensive 3D derivations we use quaternions.

\subsection{Literature Review}

The ecological approach to visual perception \cite{gibson2014ecological} established that optical flow contains rich information for navigation and spatial awareness without requiring explicit 3D reconstruction. Yet most computational methods have departed from this principle of directness. A foundational analysis of the retinal velocity field \cite{longuet1980interpretation} showed that surface structure and observer motion can be recovered from derivatives of the image flow, but this formulation, and the structure-from-motion pipelines that followed \cite{hartley2003multiple}, \cite{heeger1992subspace}, requires computing full optical flow, decomposing it into translational and rotational components, and solving for egomotion before recovering depth. Each stage introduces cost, noise sensitivity, and dependence on global constraints.

Optical flow estimation, from early variational methods \cite{horn1981determining} to modern approaches such as RAFT \cite{teed2020raft}, remains a dense, high-dimensional intermediate representation that entangles translational and rotational contributions. On the perceptual side, the optical variable tau ($\tau$) was identified as a direct estimate of time-to-contact without computing distance or speed \cite{lee1976theory}. However, time-to-contact relates to depth along the direction of motion, whereas looming relates to range, a distinction with significant geometric consequences.

Recent years have brought dramatic advances in learned 3D reconstruction. Transformer-based architectures have been applied to dense stereo reconstruction without camera calibration \cite{wang2024dust3r}. Foundation models for monocular depth estimation leverage massive datasets and Vision Transformers to achieve impressive generalization \cite{yang2024depth}, \cite{yang2024depthv2}. Real-time radiance field rendering has been enabled by 3D Gaussian Splatting \cite{kerbl20233d}, and neural scene representations continue to advance rapidly \cite{mildenhall2021nerf}. While these methods achieve remarkable results, they rely on extensive training data, significant computational resources, and learned priors. They do not provide a principled analytical relationship between visual motion cues and 3D structure, nor do they yield instantaneous per-point quantities from raw image motion.

Visual looming has been studied as a direct, measurable navigation signal \cite{Raviv2000TheVL}, and later extended to computation from the motion field and surface normals \cite{vehits23}. A visual-motion fixation invariant was identified in \cite{323828}, while spherical retinal flow for a fixating observer was analyzed in \cite{thomas1994spherical}. More recently, a closed-form vision-based solution for measuring object rotation rate by tracking a single point was presented in \cite{raviv2025visionbasedclosedformsolutionmeasuring}, providing a practical means of obtaining the perceived rotation cue $\omega$.

Despite these advances, looming and perceived rotation have largely been studied independently. To our knowledge, no prior framework explicitly combines these cues into a single analytical representation, whether classical, learning-based, or perception-driven. We have also not found in the literature an approach that unifies both cues into a single closed-form representation capable of directly yielding scaled 3D structure and heading from instantaneous per-point measurements. The \textit{OWL} function addresses this gap.
 
\section{Introducing the Complex Ratio $\tilde{t}$/$\tilde{r}$}
\subsection{General}
Let $\tilde{t}$ and $\tilde{r}$ denote two physical complex quantities in camera reference frame:
\begin{itemize}
\item $\tilde{t}$ represents the instantaneous relative translational component between the camera and a 3D point expressed as a complex number with associated units of measurement. It is measured in speed units, e.g., [meter/sec].
\item $\tilde{r}$ represents the instantaneous relative range, also expressed as a complex number, from the camera to a 3D point, with associated units of measurement. It is measured in distance units e.g., [meter].
\end{itemize}
Note that $\tilde{t}$ and $\tilde{r}$ are \textbf{not} vectors.   
To get the basic idea we can think of $\tilde{t}$/$\tilde{r}$ as a new \textbf{complex} physical quantity that represents the ratio between these two well-known \textbf{complex} physical quantities. The units of $\tilde{t}$/$\tilde{r}$ are in [1/time], for example [1/sec]. 

Although the ratio could in principle be obtained by computing $\tilde{t}$ and $\tilde{r}$ independently, doing so for every 3D point is computationally expensive. Instead our approach allows us \textbf{to mathematically derive and \textbf{directly obtain the complex $\tilde{t}$/$\tilde{r}$ }ratio from fundamental visual motion cues, the perceived visual looming ($L$) and the perceived rate of rotation ($\omega$)}, bypassing the need for any prior knowledge or computation or measurement of $\tilde{t}$ and/or $\tilde{r}$. It is obtained from one set of cue values per point and per time instant, allowing for parallelism and simplicity. 

Our discussion of $\tilde{t}$/$\tilde{r}$ explains the fundamentals of the approach, outlines some of its properties and advantages, and, most importantly, lays the groundwork for its reciprocal expression $\tilde{r}$/$\tilde{t}$, which we refer to as \textit{OWL}. We later show that for a more rigorous 3D analysis it can be expressed using quaternions.

\subsection{Understanding $\tilde{t}$/$\tilde{r}$}
As mentioned earlier $\tilde{t}$ and $\tilde{r}$ are \textbf{not} vectors. They are complex numbers with different units. However to get a basic understanding of $\tilde{t}/\tilde{r}$ we start with the vectors $\mathbf{t}$ and $\mathbf{r}$.

\begin{figure}[h]
	\centering
	{\epsfig{file = 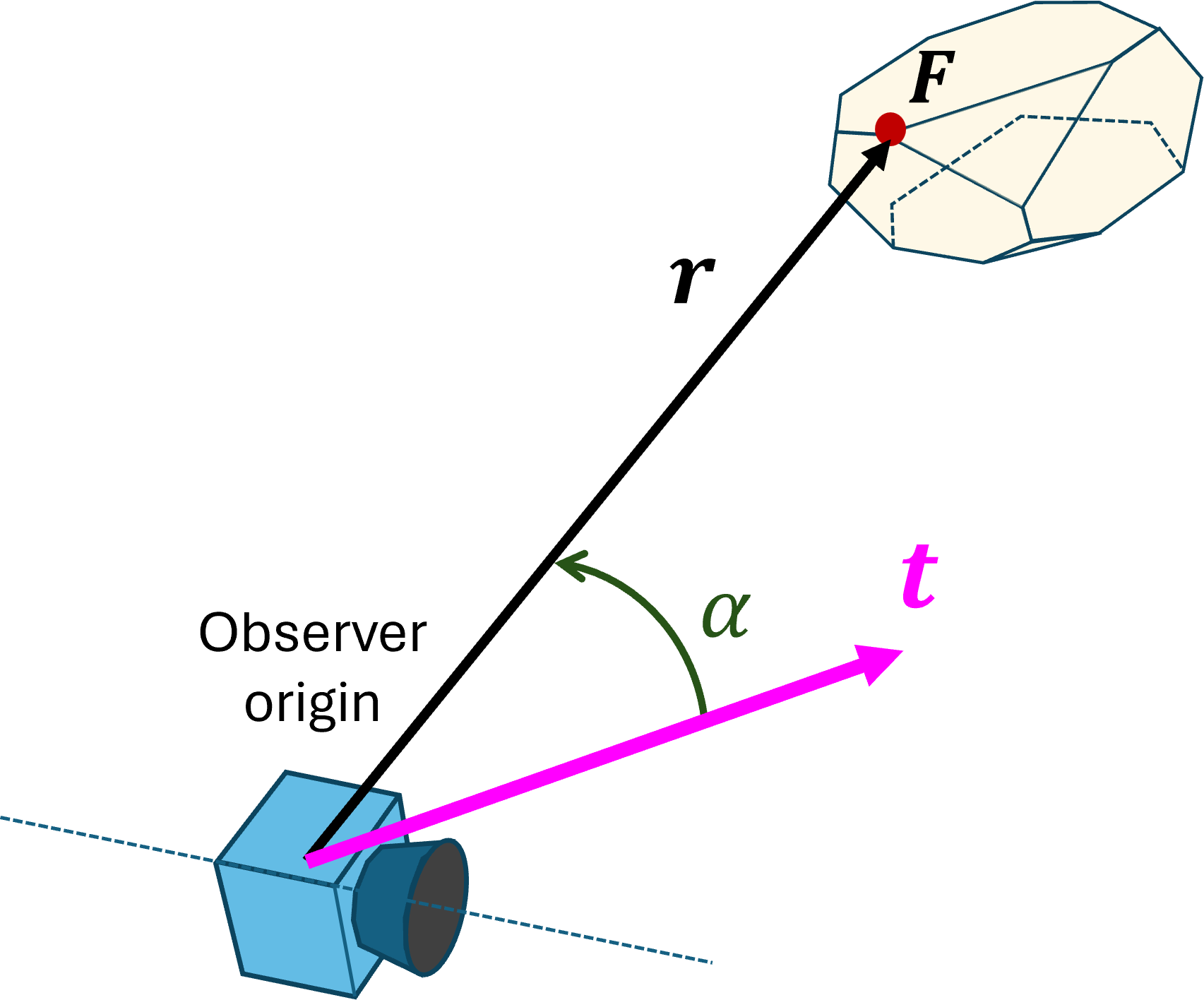, width = 8cm}}
	\caption{Motion relative to a 3D fixation point located on a rigid object}
	\label{fig001}
\end{figure}

Imagine a scenario where a camera is in relative motion with respect to a 3D object, observing a point ($F$) on the 3D object (Figure \ref{fig001}). The vector $\mathbf{t}$ represents the instantaneous relative translational velocity of the camera, while the vector $\mathbf{r}$ is the instantaneous range vector from the camera to the 3D point. The instantaneous angle $\alpha$ is measured from $\mathbf{t}$ and $\mathbf{r}$. Also let $\boldsymbol{\hat{t}}$ represent unit vector of $\boldsymbol{t}$ and $\boldsymbol{\hat{\omega}}$ represent unit vector of $\boldsymbol{\omega}$. Note that without loss of generality, any relative motion between a moving point in space and a moving camera, regardless of the camera's motion or the point's position can be analyzed, as if the camera moving relative to a stationary point $F$ in 3D (Figure \ref{fig001}). 

Now, instead of using vectors, we refer to $\tilde{t}$ and $\tilde{r}$ as two complex numbers in the plane that contains both and the angle $\alpha$. This approach allows us to take the ratio of the two complex numbers.

\paragraph*{Relating $\tilde{t}$/$\tilde{r}$ to visual cues - an intuitive introduction}

We start with two basic observations.
We based our approach on two very basic visual cues relative to a fixation point $F$: 1) Perceived looming effect from points near the fixation point, and 2) Perceived rotation of the rigid object relative to the fixation point. 

\begin{figure}[h]
	\centering
	{\epsfig{file = 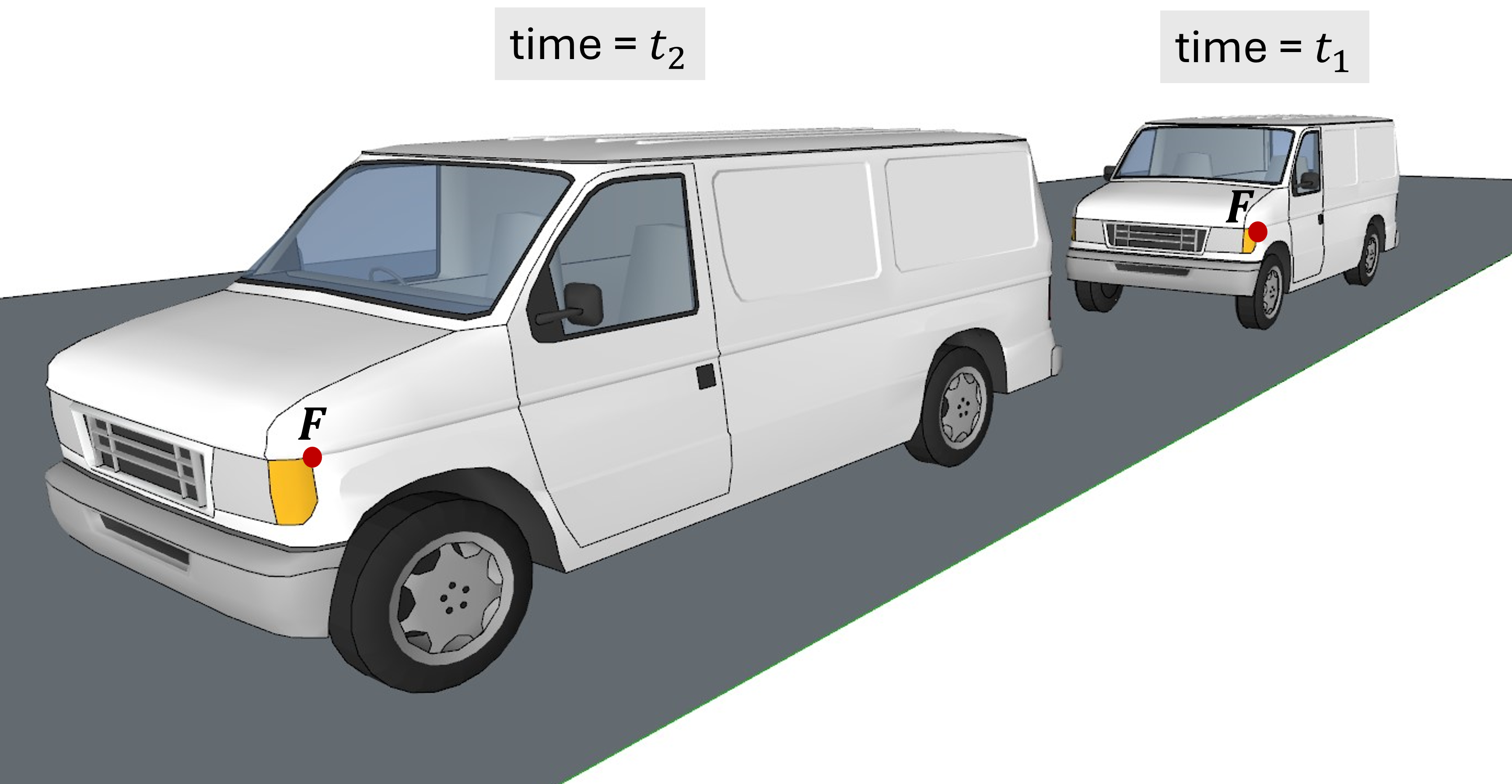, width = 8.5cm}}
	\caption{Fixating on a point $F$ of a moving rigid body at two time instants}
	\label{fig002}
\end{figure}

To clarify this observation, consider two specific time instants from a simulated image sequence capturing a car moving in rectilinear motion with no rotation (Figure \ref{fig002}). Intuitively, when fixing attention on a specific point on the car, say $F$, the surrounding region not only exhibits predominantly local mostly expansion in the image but also induces a perceived rotation around that point. (In Figure \ref{fig002} the perception is in the CW direction).

It is clear that despite these perceived cues which have different values for different time instants and for different fixation points, our overall perception of the car remains unchanged!

\begin{figure}[h]
	\centering
	{\epsfig{file = 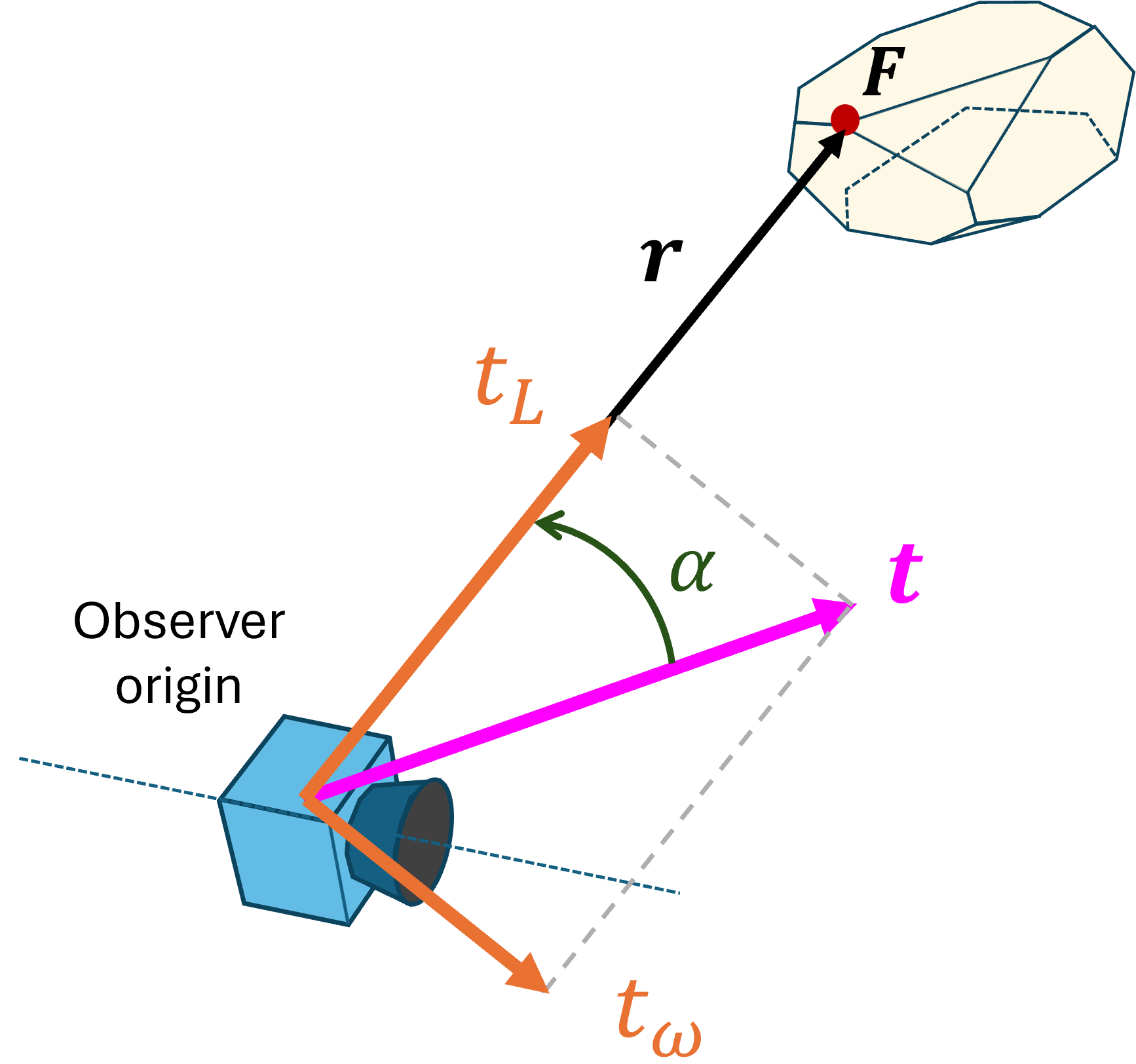, width = 8cm}}
	\caption{Relative translation decomposed into two orthogonal components}
	\label{fig003}
\end{figure}

Refer to Figure \ref{fig003}. The relative translational vector $\mathbf{t}$ in Figure \ref{fig001} can be decomposed into two orthogonal components $\mathbf{t}_{L}$ and $\mathbf{t}_{\omega}$. 

$\mathbf{t}_{L}$ is the component directed towards point $F$ and is responsible for the perceived looming effect (as perceived by the camera), which arises from the relative change in range.

$\mathbf{t}_{\omega}$ is the component perpendicular to $\mathbf{r}$, causing the camera to rotate around the fixation point in the counterclockwise (CCW) direction. In Figure \ref{fig003}, however this rotation can also be described as the camera perceiving the object as rotating around the fixation point in the opposite direction, which in this case is clockwise (CW).

The perceived looming (L) is a scalar defined as \cite{Raviv2000TheVL}   
\begin{align}
L = -\frac{\frac{dr}{dt}}{r} 
\end{align}

In our case (Figure \ref{fig003}) its value is
\begin{align}
L = \frac{|\mathbf{t}_{L}|}{|\mathbf{r}|} = \frac{|\mathbf{t}| \cos(\alpha)}{|\mathbf{r}|} \label{eq:tcos}
\end{align}

The signed magnitude of the perceived rotation is
\begin{align}
\omega = -\frac{|\mathbf{t}_{\omega}|}{|\mathbf{r}|} = -\frac{|\mathbf{t}|\sin(\alpha)}{|\mathbf{r}|} \label{eq:tsin}
\end{align}
Note that a positive value of the perceived $\omega$ corresponds to CCW direction and negative value corresponds to CW direction.

Regarding Roll of the camera: Note that we are focusing on the effect of perceived rotation due to relative instantaneous translation only, so in this case the perceived $\omega$ in \eqref{eq:tsin} excludes roll around $\mathbf{r}$.

\subsection{Unifying Fundamental Equivalence between $\tilde{t}$/$\tilde{r}$ and $\omega\&L$ Using Complex Numbers}

\begin{figure}[h]
	\centering
	{\epsfig{file = 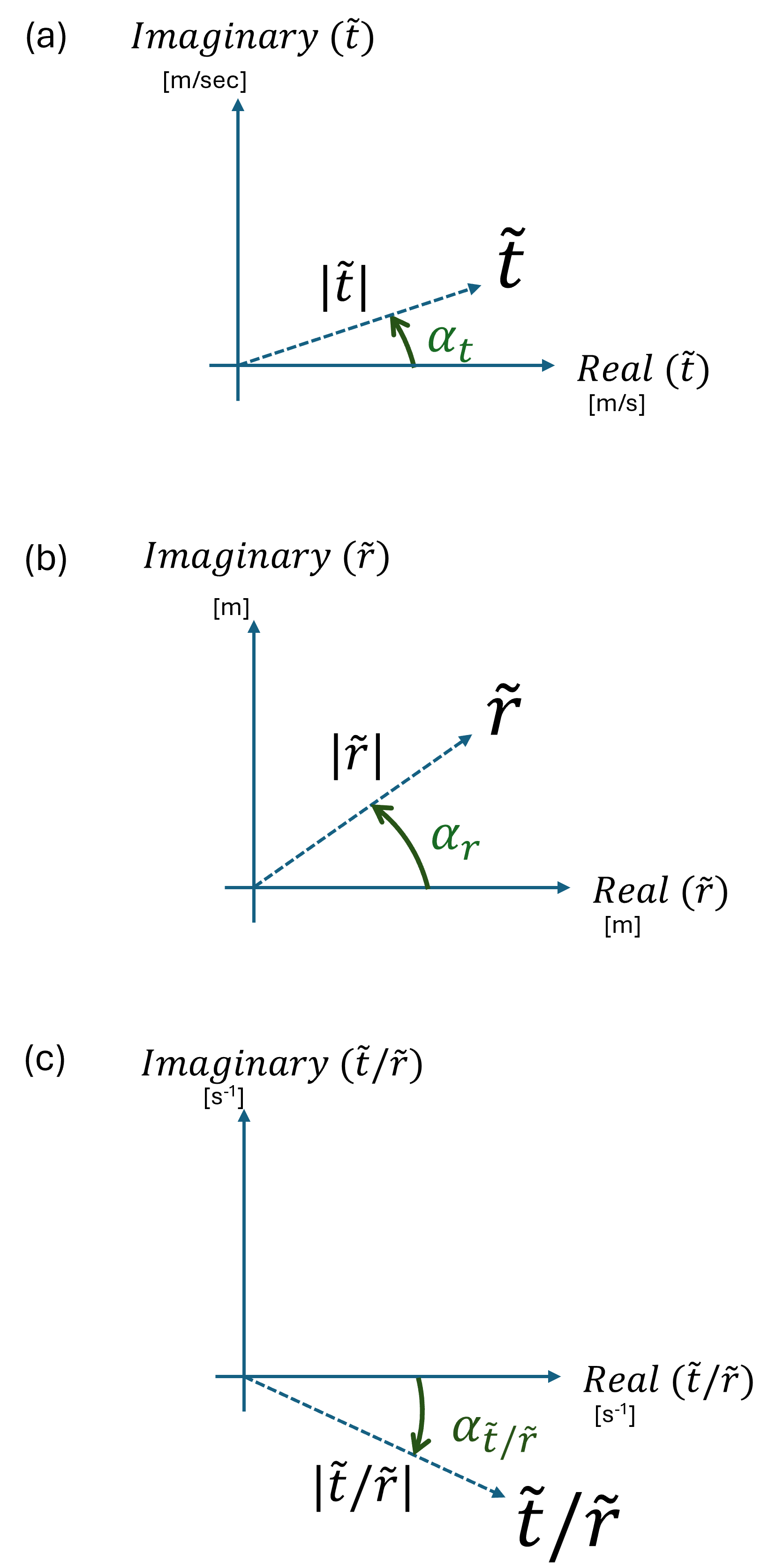, width = 6cm}}
	\caption{Complex number representation of $\tilde{t}$, $\tilde{r}$ and $\tilde{t}/\tilde{r}$}
	\label{fig0020}
\end{figure}

If the vectors $\mathbf{t}$ and $\mathbf{r}$ are represented as complex numbers $\tilde{t}$ and $\tilde{r}$, they must be plotted on separate complex planes since they have different units ([distance/time] units for $\tilde{t}$ and [distance] units for $\tilde{r}$). See Figures \ref{fig0020}(a) and \ref{fig0020}(b). 

However, the ratio of the complex entities $\tilde{t}$ and $\tilde{r}$, i.e.,  ($\tilde{t}$/$\tilde{r}$), can be plotted on another single complex plane. Specifically, dividing these two complex values each with its own units yields $\tilde{t}/\tilde{r}$, a complex value with units of [1/time]. 

Let $\alpha_{\tilde{t}/\tilde{r}}$ be 
\begin{align}
\alpha_{\tilde{t}/\tilde{r}} = \alpha_{\tilde{t}} - \alpha_{\tilde{r}} \label{eq:alpha_t_r}
\end{align}

Note that in Figure \ref{fig0020}(c) $\alpha_{\tilde{t}/\tilde{r}}$ is negative. Referring to Figure \ref{fig001} it is clear that 
\begin{align}
\alpha = - \alpha_{\tilde{t}/\tilde{r}} \label{eq:minus_alpha}
\end{align} 

The complex value of a point in the $\tilde{t}$/$\tilde{r}$ plane is obtained from:
\begin{align}
\frac{\tilde{t}}{\tilde{r}} &= \frac{|\tilde{t}|e^{j\alpha_{t}}}{|\tilde{r}|e^{j\alpha_{r}}} = \frac{|\tilde{t}|}{|\tilde{r}|}e^{j\alpha_{\tilde{t}/\tilde{r}}} = \frac{|\tilde{t}|}{|\tilde{r}|}e^{-j\alpha} \label{eq:euler1}
\end{align}

Using the Euler identity we obtain \eqref{eq:euler2} and \eqref{eq:OWL1}
\begin{align}
\frac{\tilde{t}}{\tilde{r}} &= \frac{|\tilde{t}|}{|\tilde{r}|}\cos\alpha_{\tilde{t}/\tilde{r}} + j\frac{|\tilde{t}|}{|\tilde{r}|}\sin\alpha_{\tilde{t}/\tilde{r}} \label{eq:euler2}
\end{align}

\begin{align}
\frac{\tilde{t}}{\tilde{r}} &= \frac{|\tilde{t}|}{|\tilde{r}|}\cos(-\alpha) + j\frac{|\tilde{t}|}{|\tilde{r}|}\sin(-\alpha) \label{eq:OWL1}
\end{align}

By relating equation (\ref{eq:OWL1}) to equations (\ref{eq:tcos}) and (\ref{eq:tsin}) we notice that it can be re-written as 

\setlength{\fboxrule}{2pt}  
\setlength{\fboxsep}{8pt}     
\begin{align}
\boxed{\scalebox{1.25}{$\displaystyle \frac{\tilde{t}}{\tilde{r}} = L + j\omega$}} 
\label{eq:OWL2}
\end{align}
\setlength{\fboxrule}{0.4pt}  
\setlength{\fboxsep}{4pt}     

Recall that $L$ is the signed magnitude of the Looming value, where positive Looming indicates a decreasing range, and $\omega$ is the signed magnitude of the rate of rotation, where positive $\omega$ indicates CCW direction. Both $L$ and $\omega$ are measured in [1/time] units.

Specific values of $L$ and $\omega$ are associated with a particular point on the 3D object. These values are instantaneous and vary for different 3D points.

\textbf{Equation (\ref{eq:OWL2}) shows that we can obtain directly the ratio $\tilde{t}/\tilde{r}$ from seemingly unrelated two visual motion cues, i.e., the perceived visual looming $L$ and the perceived visual rotation $\omega$, without needing specific knowledge of the range $\tilde{r}$ and/or $\tilde{t}$.}

From a practical point of view, both $L$ and $\omega$ are quantities per point per time instant directly from the image sequence raw data. They can be computed in parallel for all trackable points and their nearby points in the image. 

To further clarify the meaning of the results shown in equation (\ref{eq:OWL2}), the new combined cue-based complex domain, $\tilde{t}$/$\tilde{r}$, possesses several properties that make it highly versatile and efficient. 
\begin{enumerate}[label={(\arabic*)}, leftmargin=*]
\item Both $L$ and $\omega$ depend solely on the relative instantaneous translation vector between the camera and a specific point in 3D, remaining unaffected by camera rotation. 
\item The $\tilde{t}$/$\tilde{r}$ cue is environment-independent, meaning it applies to any point in 3D space which is moving relative to the camera, and \item $\tilde{t}$/$\tilde{r}$ is calculated individually for each point allowing parallel processing. 
\end{enumerate}
This includes independently moving points, such as those on different rigid bodies.

\paragraph*{Two Observations About $\omega$ and $L$}
In the following two sections we show two important relations between $L$ and $\omega$ and $\tilde{r}$ and $\tilde{t}$ for each specific 3D point. The first is related to the relative magnitude of the range $\tilde{r}$ (i.e., range up to a scale factor) and the second is related to relative direction of motion $\boldsymbol{\hat{t}}$ 

\subsection{Obtaining Scaled 3D from $\omega$ and $L$}

\setlength{\fboxrule}{1pt}  
\setlength{\fboxsep}{8pt} 
Using equation \eqref{eq:OWL2}
\begin{align}
\boxed{\frac{|\tilde{t}|}{|\tilde{r}|} = \sqrt{L^2 + |{\omega}|^2} \label{eq:wlsquare}}
\end{align}
\setlength{\fboxrule}{0.4pt}  
\setlength{\fboxsep}{4pt}     

This implies that, given $\omega$ and $L$, the relative distance, i.e., the magnitude of range of any 3D point relative to the camera under relative translational motion can be obtained up to an unknown relative speed scale factor \(\mathbf{|t|}\). \textbf{Accordingly, by fusing $\omega$ and $L$ as demonstrated, it is possible to reconstruct the scene structure up to this scale factor $\mathbf{|t|}$}.

The scaled reconstruction will be further explained when we discuss the reciprocal of $\left(\tilde{t}/\tilde{r}\right)$, i.e., \(OWL = \left( \tilde{r}/\tilde{t}\right)\).

Assuming that $\omega$ and $L$ (including their sign) are available, these results hold for any relative translational motion and any 3D point, including points that belong to different moving non-rotating rigid bodies.

\subsection{Obtaining Direction of Motion with $\omega$ and $L$}
Refer to Figure \ref{fig0020} and Equations \eqref{eq:OWL1} and \eqref{eq:OWL2}. Clearly

\setlength{\fboxrule}{1pt}  
\setlength{\fboxsep}{8pt} 
\begin{align}
\boxed{\tan(\alpha_{\tilde{t}/\tilde{r}}) = -\tan(\alpha) = - \frac{\omega}{L}}\label{eq:megainvariant}
\end{align}
\setlength{\fboxrule}{0.4pt}  
\setlength{\fboxsep}{4pt} 

This means that, using this ratio, we can compute the angle between the instantaneous relative direction of motion $\mathbf{\hat{t}}$ and the direction toward any 3D point $\mathbf{\hat{r}}$ both in camera coordinates. This can also be done in the complex domain. 

Practically, to pinpoint $\hat{\mathbf{t}}$ we need to calculate angles of at least three points. The reason, we need to intersect three cones (one per point). This intersection yields $\hat{\mathbf{t}}$.

\subsection{Summary of Major Findings}
\begin{enumerate}[label={(\arabic*)}, leftmargin=*]
\item\textbf{Equation \eqref{eq:OWL2} is the essential insight that forms the foundation of our perception-based approach. Simply put, using per point values of two visual motion cues, perceived looming ($L$) and perceived rotation ($\omega$) that can be obtained from raw data of 2D image sequence (due to relative motion) the amplitude of the instantaneous range per point can be obtained up to a speed scale factor! This scaled 3D reconstruction requires no prior knowledge.}
\item\textbf{Equation \eqref{eq:megainvariant} shows that using the ratio $\omega/L$ of a few points the instantaneous translation direction (heading) can also be obtained.}
\end{enumerate}

\subsection{Independence Properties}
The perceived looming $L$ and perceived rotation $\omega$ can be well estimated from image sequences independently of the viewing configuration. Their magnitude and sign are preserved regardless of screen size, viewing distance, screen orientation, or field of view. Consequently, these cues remain nearly unchanged when the same image sequence is observed from different display conditions.

\begin{figure}[h] 
    \centering {\epsfig{file = 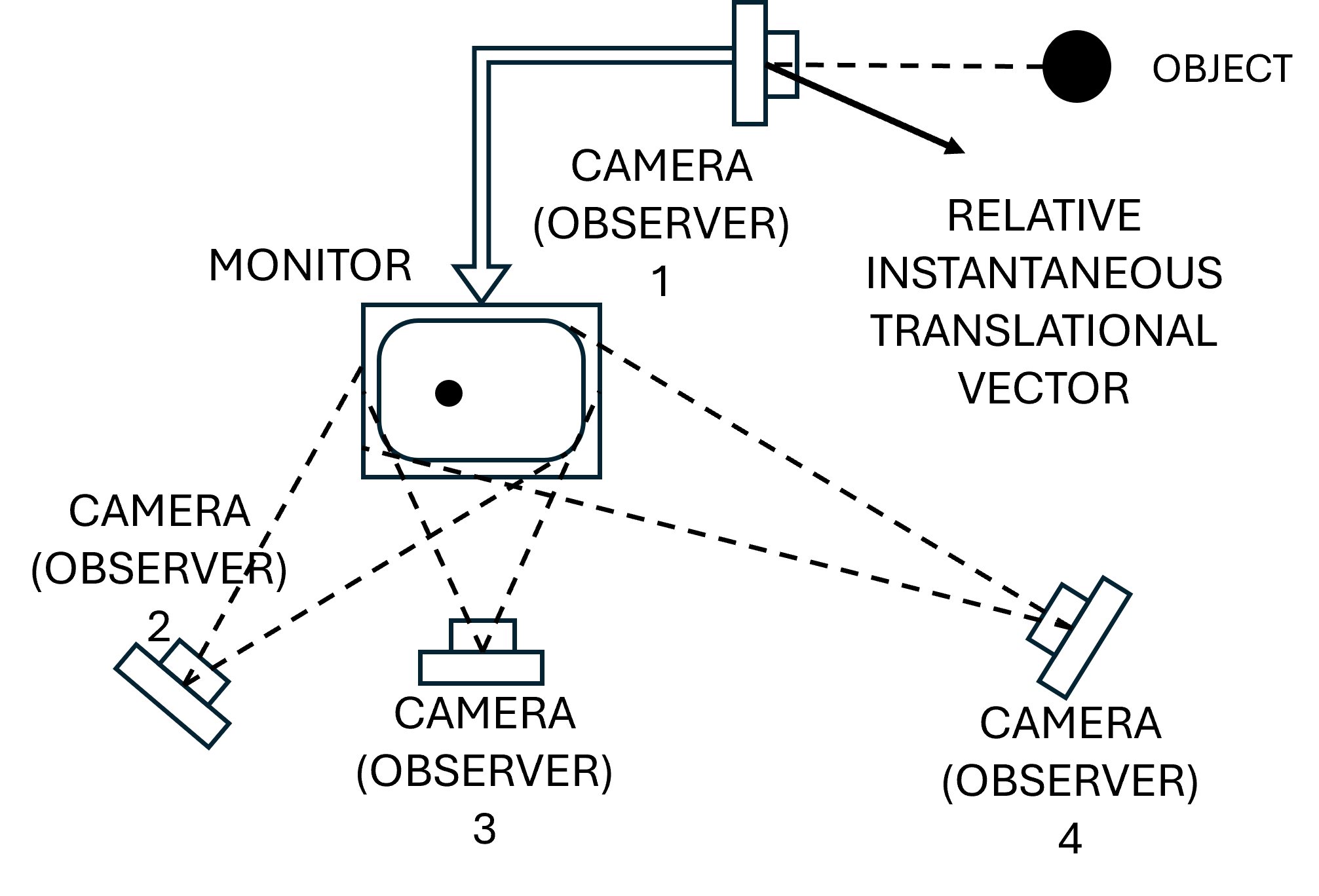, width = 7cm}} \caption{On the perception of L and w} \label{fig0021} 
\end{figure}

Figure \ref{fig0021} illustrates this property: cameras placed at different locations and orientations (1–4) observing the same scene produce nearly identical values of $L$ and $\omega$. This invariance arises because both cues depend only on relative changes in the projected image sequence rather than on absolute scene depth or camera calibration.

Since the ratio $\tilde{t}/\tilde{r}$ is derived directly from these cues, it can also be estimated without direct observation of the 3D scene. As a result, the method does not require stereo cameras, camera calibration, or detailed knowledge of camera parameters, and it can operate with a wide range of imaging systems.

\begin{figure}[h]
\centering
{\epsfig{file = 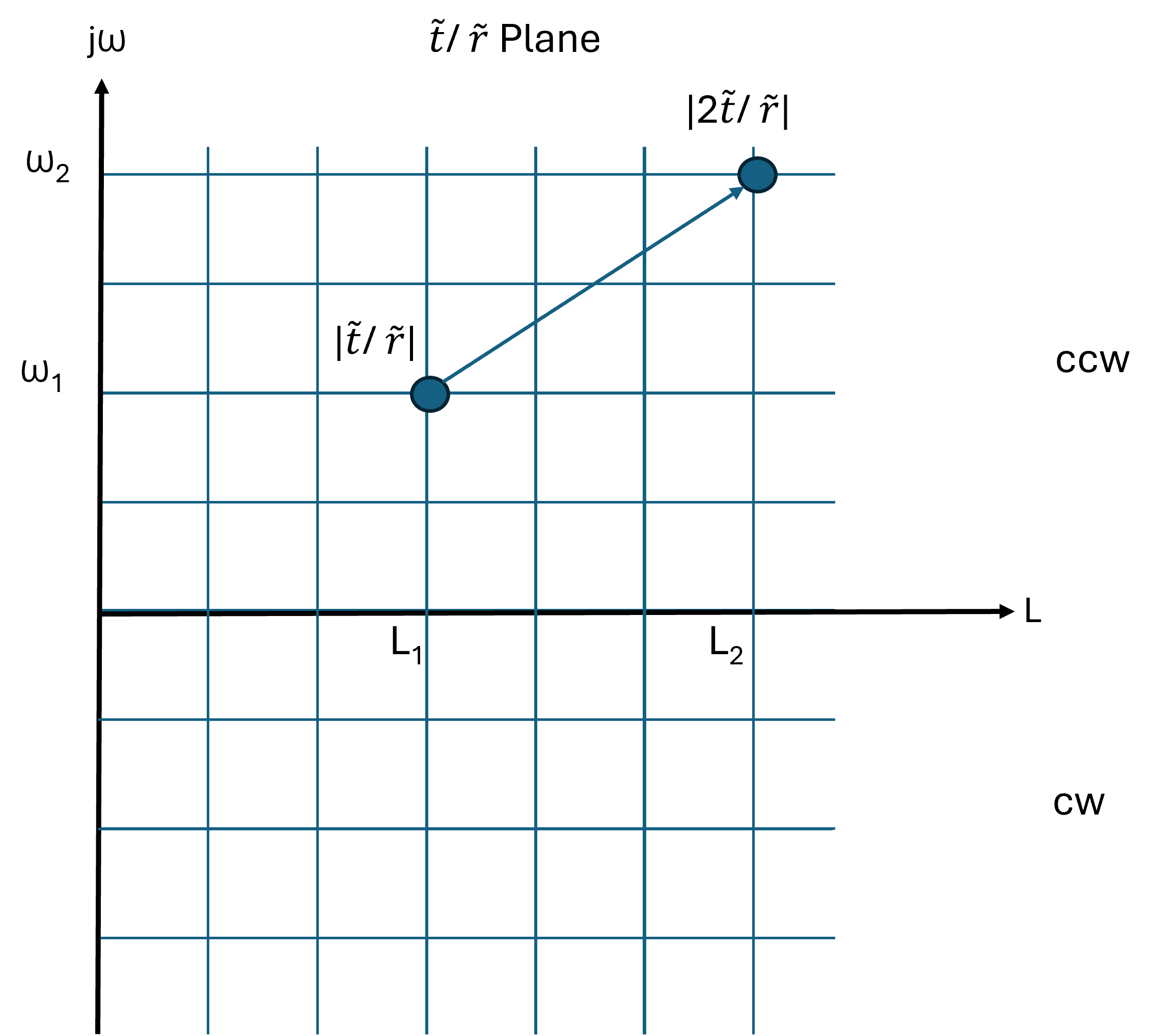, width = 8cm}}
\caption{Effect of speed on the complex value $\tilde{t}/\tilde{r}$. Increasing $|\tilde{t}|$ while keeping the motion direction constant shifts the point radially in the $\tilde{t}/\tilde{r}$ plane, scaling both $L$ and $\omega$.}
\label{fig_scaling}
\end{figure}

\paragraph{Scaling property.}
If the magnitude of the relative translation $|\tilde{t}|$ increases while the direction of motion remains unchanged, the values of $L$ and $\omega$ scale proportionally. Consequently, the complex value $\tilde{t}/\tilde{r}=L+j\omega$ moves along a radial line in the $\tilde{t}/\tilde{r}$ plane. This reflects the intuitive fact that faster motion produces proportionally stronger looming and perceived rotation, as illustrated in Fig.~\ref{fig_scaling}.

\subsection{Clarifying the Difference between Looming and TTC}
\begin{figure}[h]
	\centering
	{\epsfig{file = 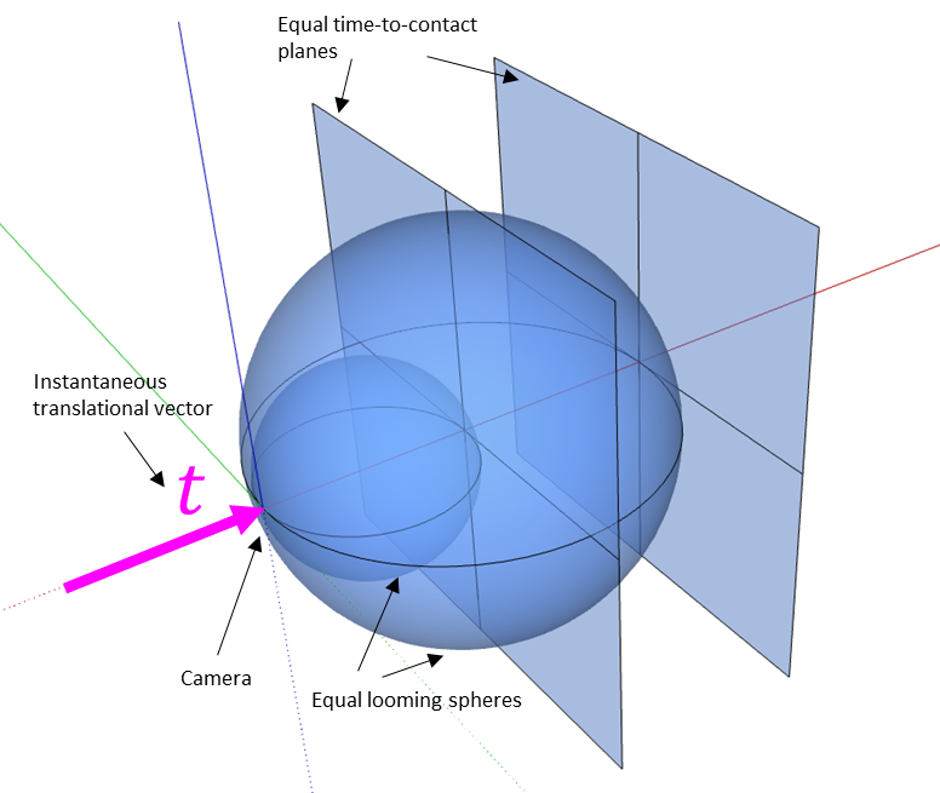, width = 8cm}}
	\caption{Equal time-to-contact (\textit{TTC}) planes and equal looming ($L$) spheres}
	\label{fig0028}
\end{figure}

Refer to Figure \ref{fig0028}. Time-to-contact (\textit{TTC}) also known as time-to-collision \cite{lee1976theory} concerns depth, whereas looming \cite{raviv1992quantitative} relates to range. Surfaces of constant \textit{TTC} are planes perpendicular to the instantaneous direction of motion. Surfaces of constant looming are spheres that lie in front of the camera, with their centers located along the instantaneous translational motion vector \cite{Raviv2000TheVL}\cite{vehits23}.

\subsection{On the relationship between $\omega$, optical flow $\dot{\theta}$, and camera rotation \(\Omega\)}
Notably, $\omega$ is distinct from optical flow, though related. Optical flow represents the total projected velocity vector in the image, arising from all motions (of both the camera and the point). It describes the change in a point’s projected location within the image. In contrast, $\omega$ represents the perceived rotation of neighboring points on the same rigid object around the point of interest. This result can be derived for the general 3D case by expressing the perceived angular velocity vector $\boldsymbol{\omega}$ and the actual camera rotation vector $\boldsymbol{\Omega}$ in spherical coordinates (see Appendix A), yielding
\begin{align}
    \omega_\phi + \Omega_\phi &= -\dot{\theta}\cos\phi \\
    \omega_\theta + \Omega_\theta &= \dot{\phi}
\end{align}
The 2D case follows as a special case of this formulation. For planar motion, the elevation angle satisfies $\phi = 0$, hence $\dot{\phi} = 0$ and $\cos\phi = 1$. Substituting into the above equations reduces the system to

\setlength{\fboxrule}{1pt}  
\setlength{\fboxsep}{8pt} 
\begin{align}
    \boxed{\omega + \Omega + \dot{\theta} = 0}
\end{align}
\setlength{\fboxrule}{0.4pt}  
\setlength{\fboxsep}{4pt} 

Based on this invariant relationship, the value of $\omega$ can be determined as the negative of the optical flow after removing the influence of \(\Omega\) (camera rotation) from it. In other words, the value of $\omega$ is the de-rotated flow with a negative sign $\omega=-(\dot{\theta} + \Omega)$. If the camera rotation and optical flow values are known, $\omega$ can be calculated at each and every point directly from the image sequence without needing observations of $\omega$ using the neighboring points.

\section{The \textit{OWL} function}

\subsection{General}
Simply put, the function \textit{OWL} is the reciprocal of $\tilde{t}$/$\tilde{r}$, i.e., $\tilde{r}$/$\tilde{t}$. We use \textit{OWL} because it offers several advantages in terms of perception and action. It stands for ``Orthogonal, $\omega$, $L$". 

In the \textit{OWL} domain, by observing many \textit{OWL} values of points stationary objects appear continuously geometrically unchanged over time, up to a speed scale factor, i.e., \textit{maintaining shape constancy despite the constantly changing large volume of dynamic visual data}. This facilitates a simple and efficient scaled reconstruction of the scene during relative motion.  Additionally, by leveraging basic visual perception cues, the \textit{OWL} representation can suggest potential camera actions (but it is beyond the scope of this paper).

\subsection{\textit{OWL} as a Reciprocal of $\tilde{t}$/$\tilde{r}$}

\begin{figure}[h]
	\centering
	{\epsfig{file = 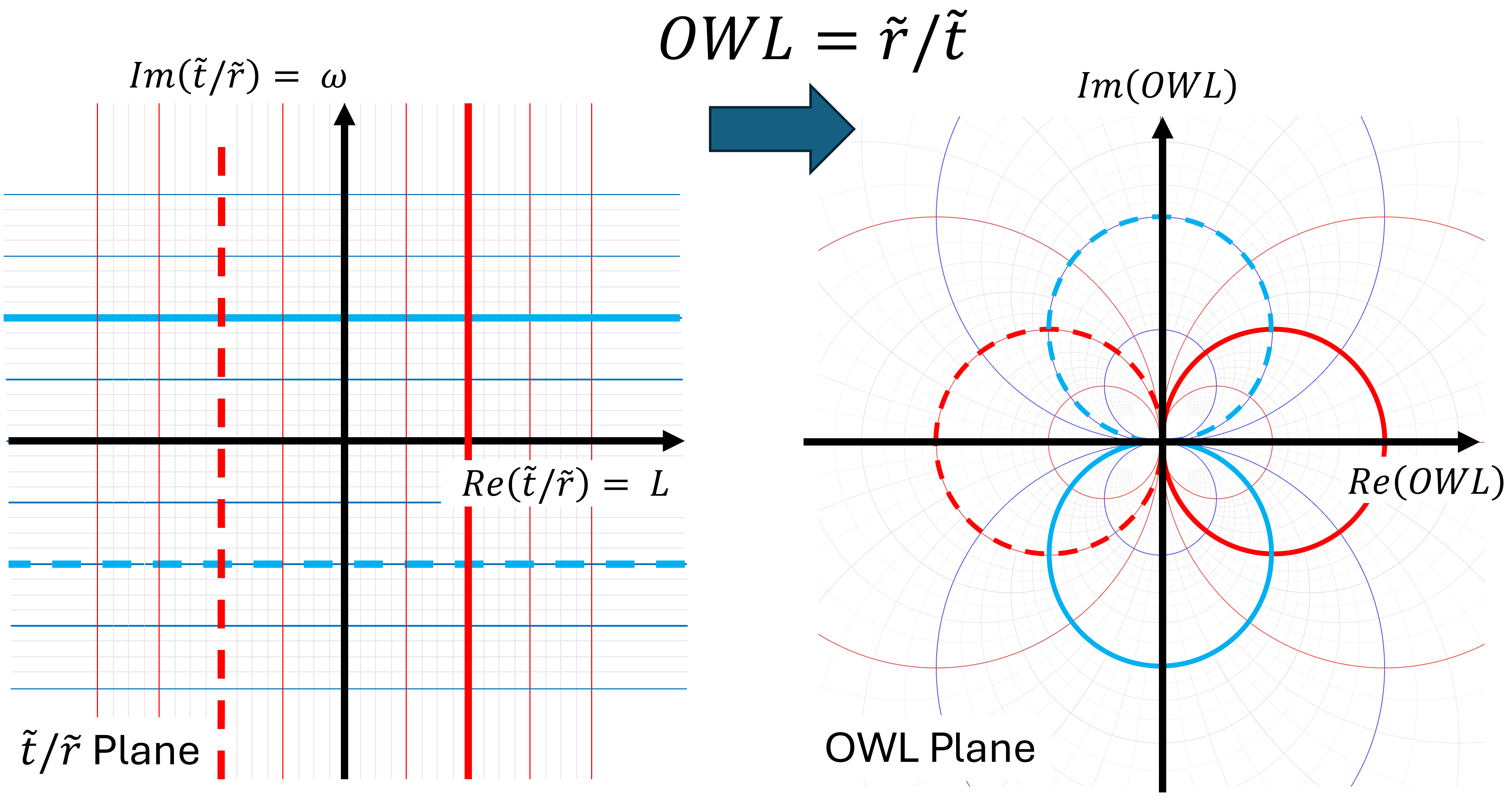, width = 8cm}}
	\caption{From $\tilde{t}$/$\tilde{r}$ lines to $OWL = \tilde{r}/\tilde{t}$ circles}
	\label{fig0029}
\end{figure}

Refer to Figure \ref{fig0029}. The \textit{OWL} function is obtained using a basic conformal mapping \cite{churchill2013complex} that transforms circles and lines from \(\tilde{t}/\tilde{r}\) in the complex plane into either circles or straight lines in the $\tilde{r}/\tilde{t}$ (\textit{OWL}) complex plane. This inversion causes points near the origin to be mapped far away, while points farther from the origin are brought closer. It is also a transformation that preserves local angles.

\subsection{Visualizing \textit{OWL}}
\begin{figure}[h]
	\centering
	{\epsfig{file = 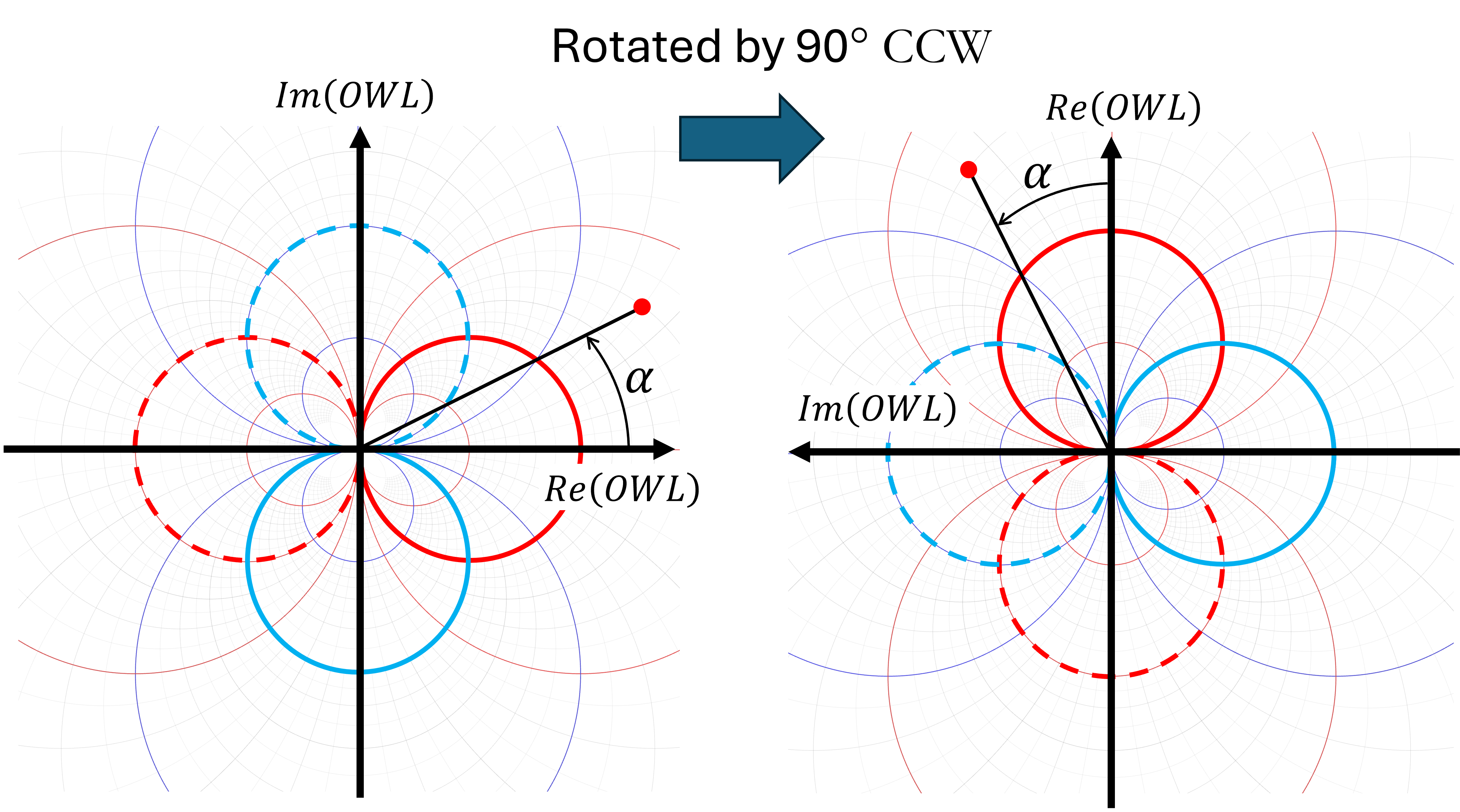, width = 8cm}}
	\caption{Rotating \textit{OWL} domain for better intuition}
	\label{fig0030}
\end{figure}

For visualization purposes we rotate \(\tilde{r}/\tilde{t}\) CCW by 90 degrees. This rotation ensures that the direction of motion is ``up" (vertical) making it more intuitive (Figure \ref{fig0030}).

\subsection{Visualizing \textit{OWL} in 3D}
\begin{figure}[h]
	\centering
	{\epsfig{file = 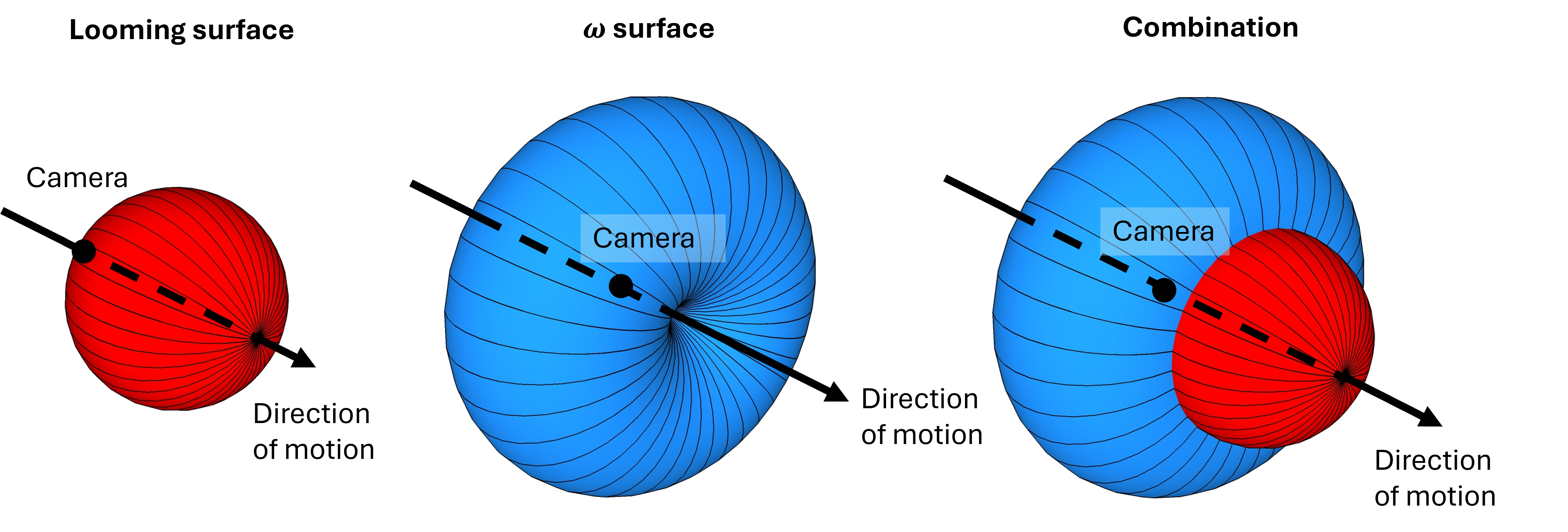, width = 8cm}}
	\caption{Visualizing an equal looming sphere and an equal $\omega$ torus}
	\label{fig0035}
\end{figure}

\textit{OWL} can be visualized in 3D through two geometric primitives: Looming ($L$) spheres and perceived rotation ($\boldsymbol{\omega}$) tori. Figure~\ref{fig0035} depicts a constant-$L$ sphere, a constant-$\boldsymbol{\omega}$ torus, and both combined, where all points on each surface share the same respective magnitude.

\subsection{\textit{OWL} of Multiple Points Potential Ambiguity}
\textit{OWL} is a function with time units, because it is related to dividing distance units by speed units. Any value of the magnitude of \textit{OWL} can be interpreted as the magnitude of range scaled by a speed factor.

This leads to potential ambiguity. For example, two different points at 5m and 10m along the same radial line from the origin, moving in the same direction at 20m/s and 40m/s respectively, will produce the same magnitude of \textit{OWL} (in this example 5/20 = 10/40 = 0.25sec).

If an exact 3D reconstruction is needed then the magnitude of the relative speed needs to be known.

\begin{figure}[h]
	\centering
	{\epsfig{file = 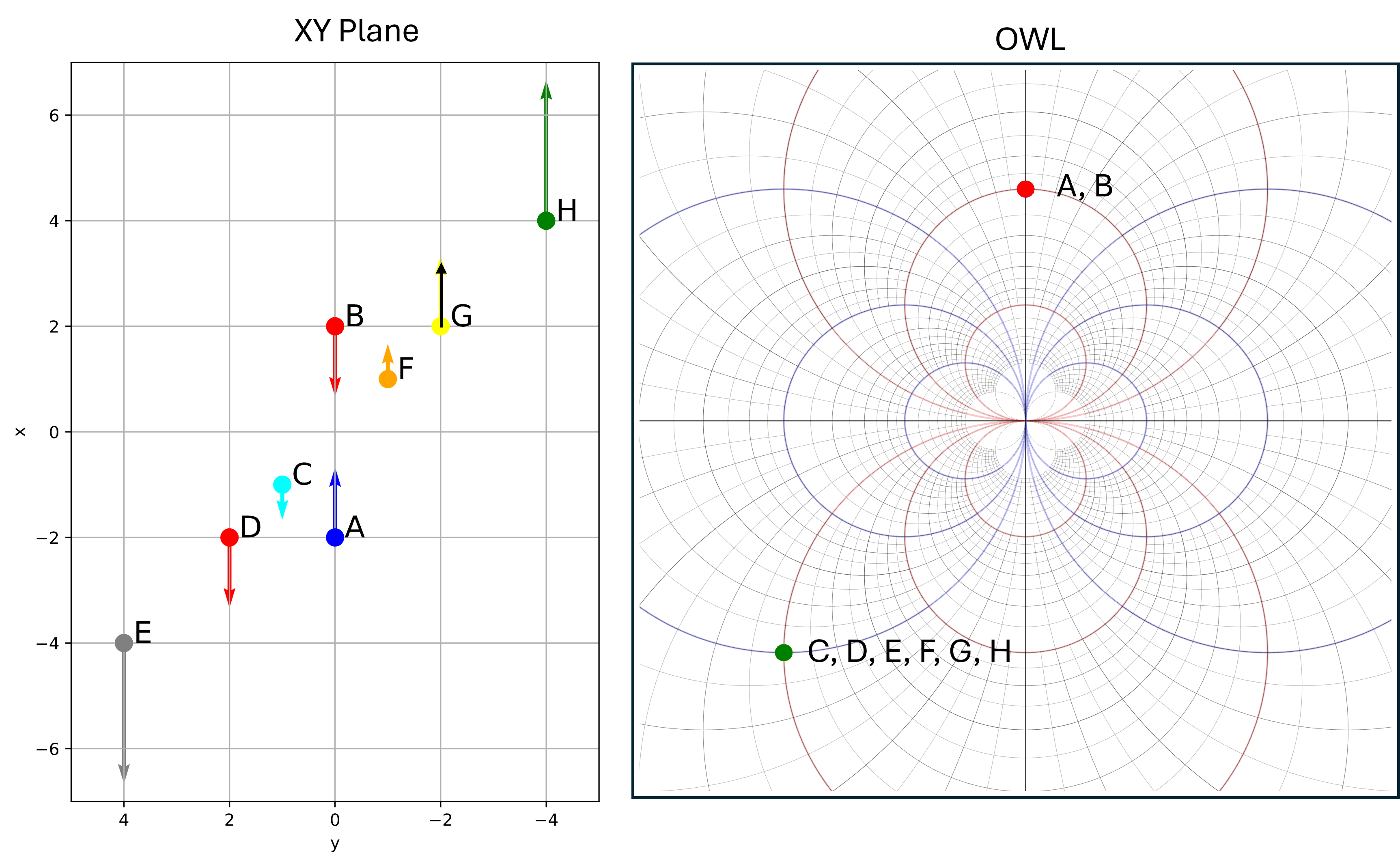, width = 8.5cm}}
	\caption{Multiple points producing identical \textit{OWL} values}
	\label{fig0041}
\end{figure}

Figure \ref{fig0041} shows examples of eight moving points over time in XY Cartesian plane as shown in Figure \ref{fig0041} (left image) where arrows refer to directions of motion and their magnitudes refer to their speeds, and the way they appear in \textit{OWL} complex domain as shown in Figure \ref{fig0041} (right image). In XY plane the motions are relative to the origin (0,0). Note the many-to-one values of the \textit{OWL} function. It shows that points with the same ratio of distance to speed in complex values appear at the same location in \textit{OWL}, specifically A and B appear at a specific point in \textit{OWL} domain, and also C,D,E,F,G and H appear at another specific point in \textit{OWL} domain.

\subsection{\textit{OWL} of Points Over Time}
\begin{figure}[htbp]
    \centering
    \includegraphics[width=0.45\textwidth]{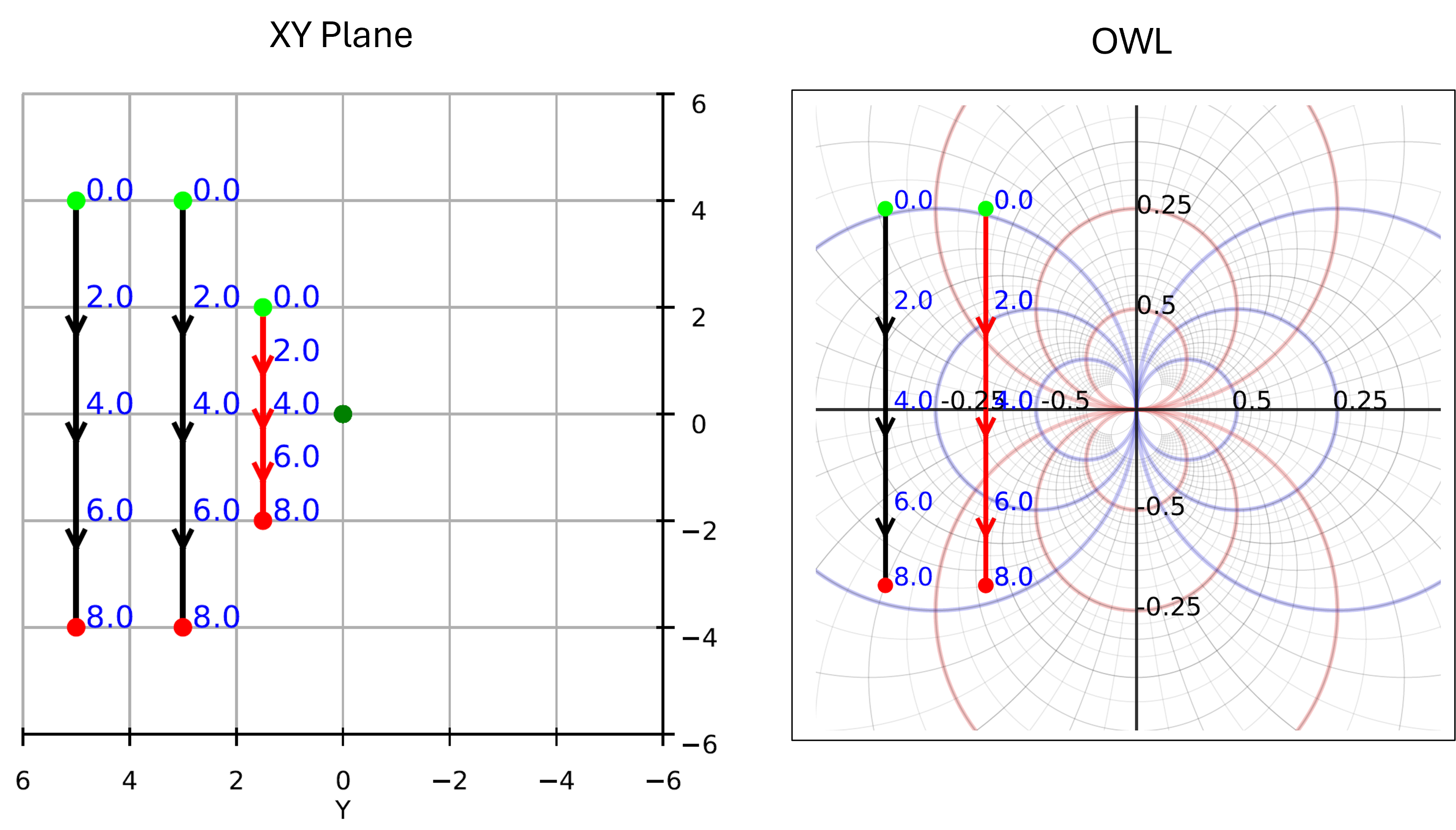}
    \caption{Rectilinear motion}
    \label{fig0043a}
\end{figure}

\begin{figure}[htbp]
    \centering
    \includegraphics[width=0.45\textwidth]{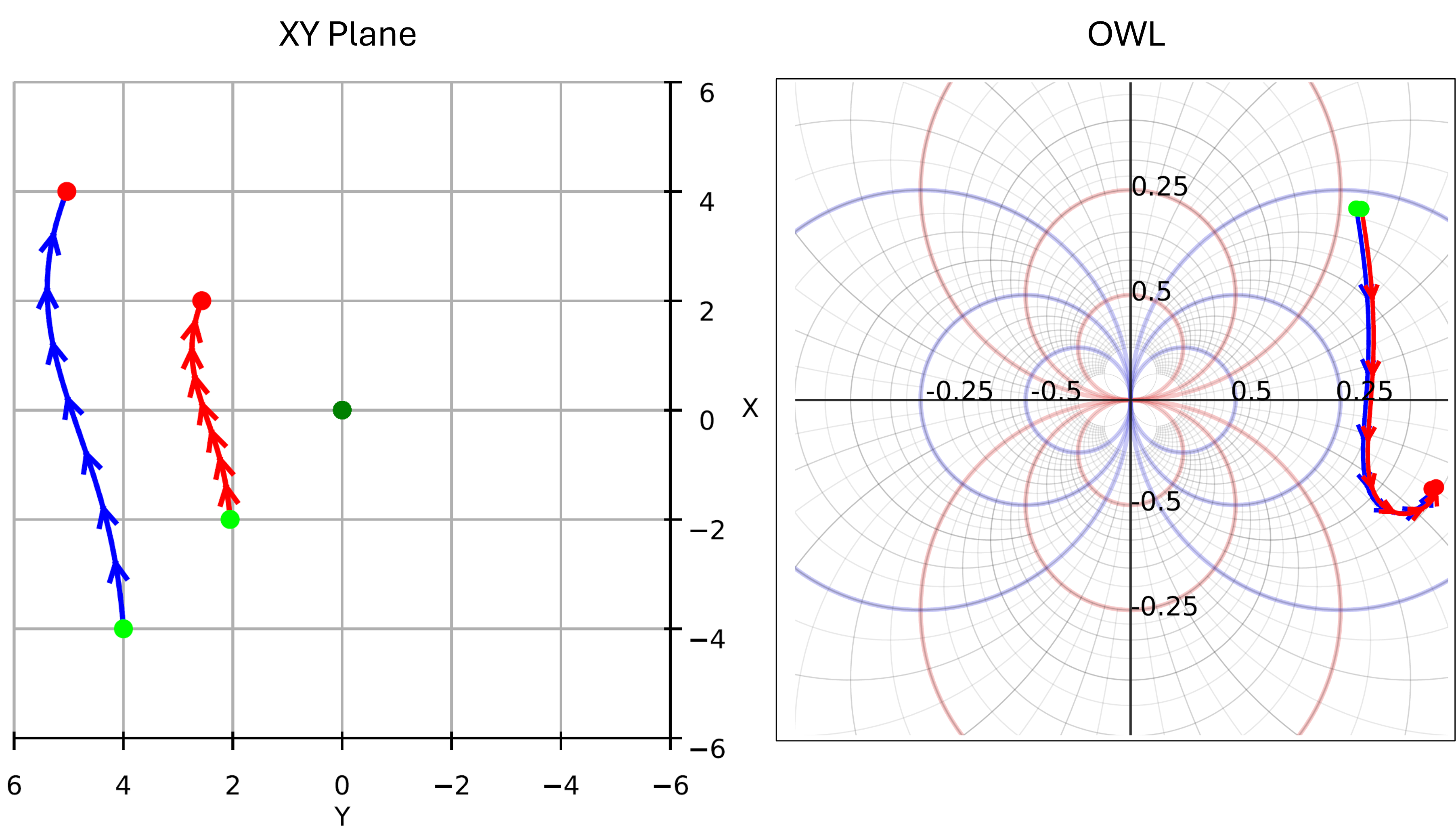}
    \caption{Scaled motion}
    \label{fig0043d}
\end{figure}

\begin{figure}[htbp]
    \centering
    \includegraphics[width=0.45\textwidth]{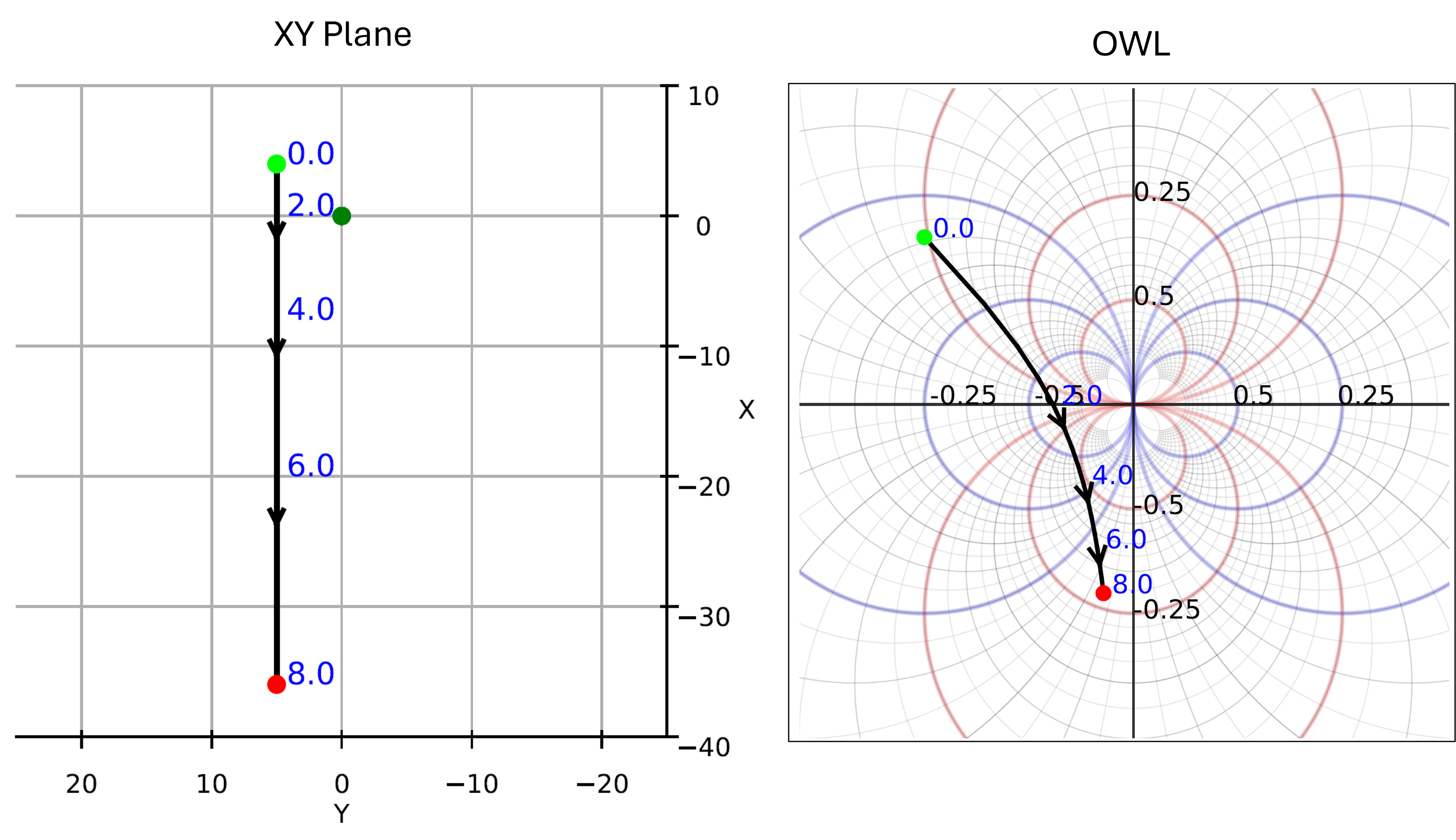}
    \caption{Constant Acceleration}
    \label{fig0043b}
\end{figure}

\begin{figure}[htbp]
    \centering
    \includegraphics[width=0.45\textwidth]{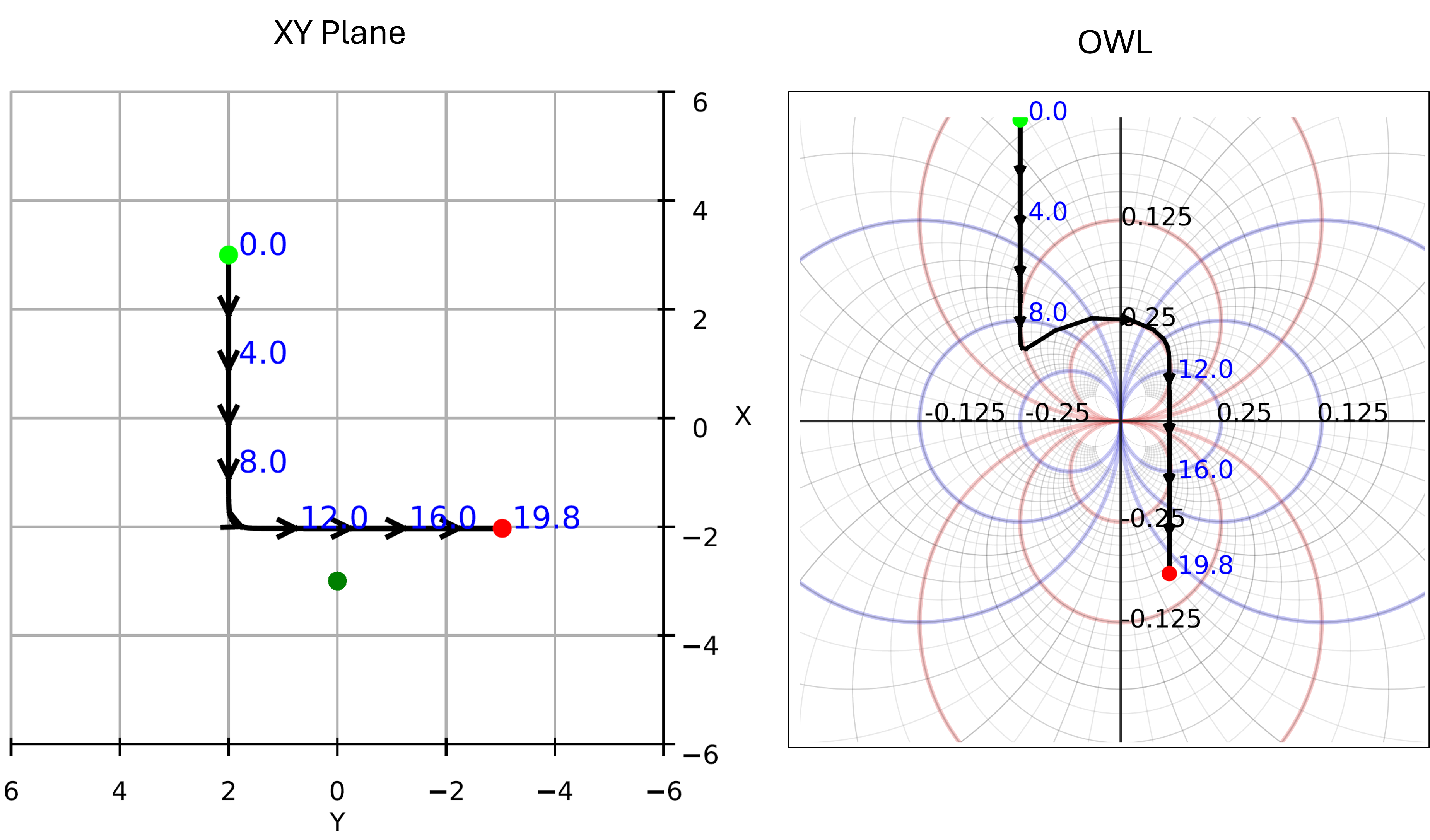}
    \caption{Change in Direction}
    \label{fig0043c}
\end{figure}

The following are 2D visual examples that illustrate relations between trajectories of points in the XY plane and the corresponding trajectories in \textit{OWL}. Using quaternions we can extend them to 3D as well. Refer to figures \ref{fig0043a} to  \ref{fig0043c}.

Figure \ref{fig0043d} shows the effect of a two different trajectories that are scaled and their identical location in \textit{OWL}. 

Figure \ref{fig0043a} (left image) shows three moving points in XY over time and the corresponding location of these points in \textit{OWL}. Note that the moving red point and the black moving (center) point appear in the same locations in \textit{OWL} due to the fact that the ratio between range and speed is the same for both.

Figure \ref{fig0043b} shows the effect of a constant acceleration point in XY and its location in \textit{OWL}. 

Figure \ref{fig0043c} shows another motion scenario.

\section{Extending the \textit{OWL} Function to 3D Using Quaternions}

Building on the 2D derivation of \textit{OWL} using complex numbers, we now extend the framework to 3D by representing the translation and range vectors as pure quaternions. Let:
\begin{align}
    R &= (0, \mathbf{r}) \label{eq:pureQ-r}\\
    T &= (0, \mathbf{t}) \label{eq:pureQ-t}
\end{align}

where $R$ and $T$ are pure quaternions, meaning their scalar part is zero and their vector parts are $\mathbf{r}$ and $\mathbf{t}$, respectively. The quaternion ratio $ToR$ is then defined as the quaternion product:
\begin{align}
    ToR &= T \otimes R^{-1} \label{eq:quaternion_product}
\end{align}

where $\otimes$ denotes the quaternion product. The reciprocal of $ToR$ is defined as:
\begin{align}
    RoT &= (ToR)^{-1} \label{eq:RoT_inverse}
\end{align}

Expressing $ToR$ as a function of $L$ and $\boldsymbol{\omega}$ yields:
\begin{align}
    ToR &= L + \boldsymbol{\omega} \label{eq:L_plus_w}
\end{align}

\paragraph{Direction of Motion}
From $ToR$, the direction of motion $\hat{\mathbf{t}}$ can be expressed as (refer to Appendix B):
\begin{align}
\hat{\mathbf{t}} &= \frac{L}{\sqrt{L^2 + \|\boldsymbol{\omega}\|^2}}\mathbf{e}_r + \frac{\boldsymbol{\omega} \times \mathbf{e}_r}{\sqrt{L^2 + \|\boldsymbol{\omega}\|^2}}
\end{align}

\paragraph{Summary of \textit{OWL} in quaternion form}
Starting from $RoT = ToR^{-1}$ and substituting $ToR = L + \boldsymbol{\omega}$, we obtain:
\begin{align}
    RoT &= {ToR}^{-1}\\
    RoT &= (L + \boldsymbol{\omega})^{-1}\\
    RoT &= \frac{ToR^*}{\|ToR\|^2}\\
    RoT &= \frac{L - \boldsymbol{\omega}}{L^2 + \|\boldsymbol{\omega}\|^2}\\
    RoT &= \frac{L}{L^2 + \|\boldsymbol{\omega}\|^2} - \frac{\boldsymbol{\omega}}{L^2 + \|\boldsymbol{\omega}\|^2} \label{eq:RoT_inw_L}
\end{align}

\section{Initial Simulation Results}

As mentioned earlier, when referring to the 3D formulation we use quaternions and the quantity $RoT$ instead of the complex-domain OWL representation. Two simulation experiments were conducted to validate the analytical framework.

\begin{figure}[ht]
    \centering
    \includegraphics[width=0.4\textwidth]{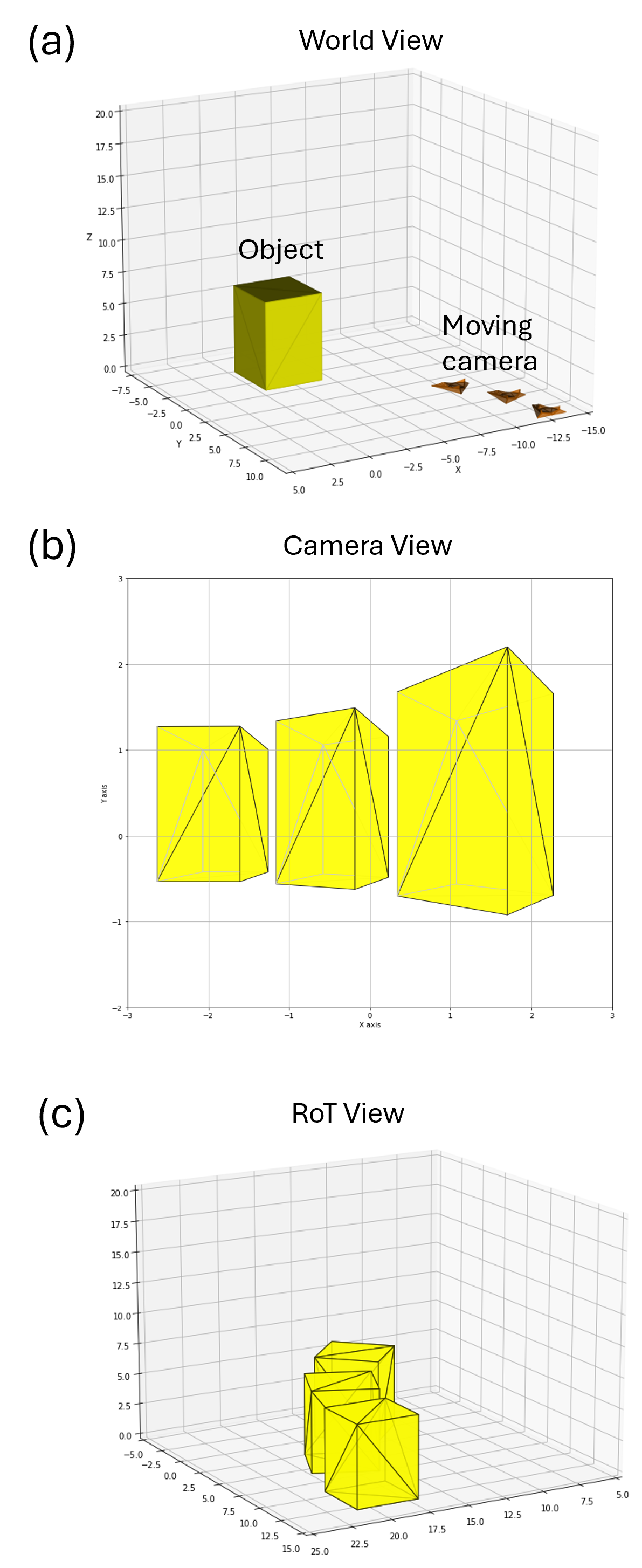}
    \caption{Reconstruction of a rigid object from OWL (RoT) values over time as the camera moves}
    \label{fig0052b}
\end{figure}

The first experiment (Figure~\ref{fig0052b}) is a Python-based simulation used to verify the analytical pipeline. A stationary rigid object (a cube) is observed by a camera undergoing translational motion. The simulation shows (a) the world configuration with the moving camera and the object, (b) the sequence of images as observed by the camera over several frames, and (c) the reconstruction of the object in the $RoT$ domain. Despite the continuously changing image projections, the object remains geometrically consistent in the $RoT$ representation, demonstrating the predicted constancy property.

\begin{figure}[ht]
    \centering
    \includegraphics[width=0.48\textwidth]{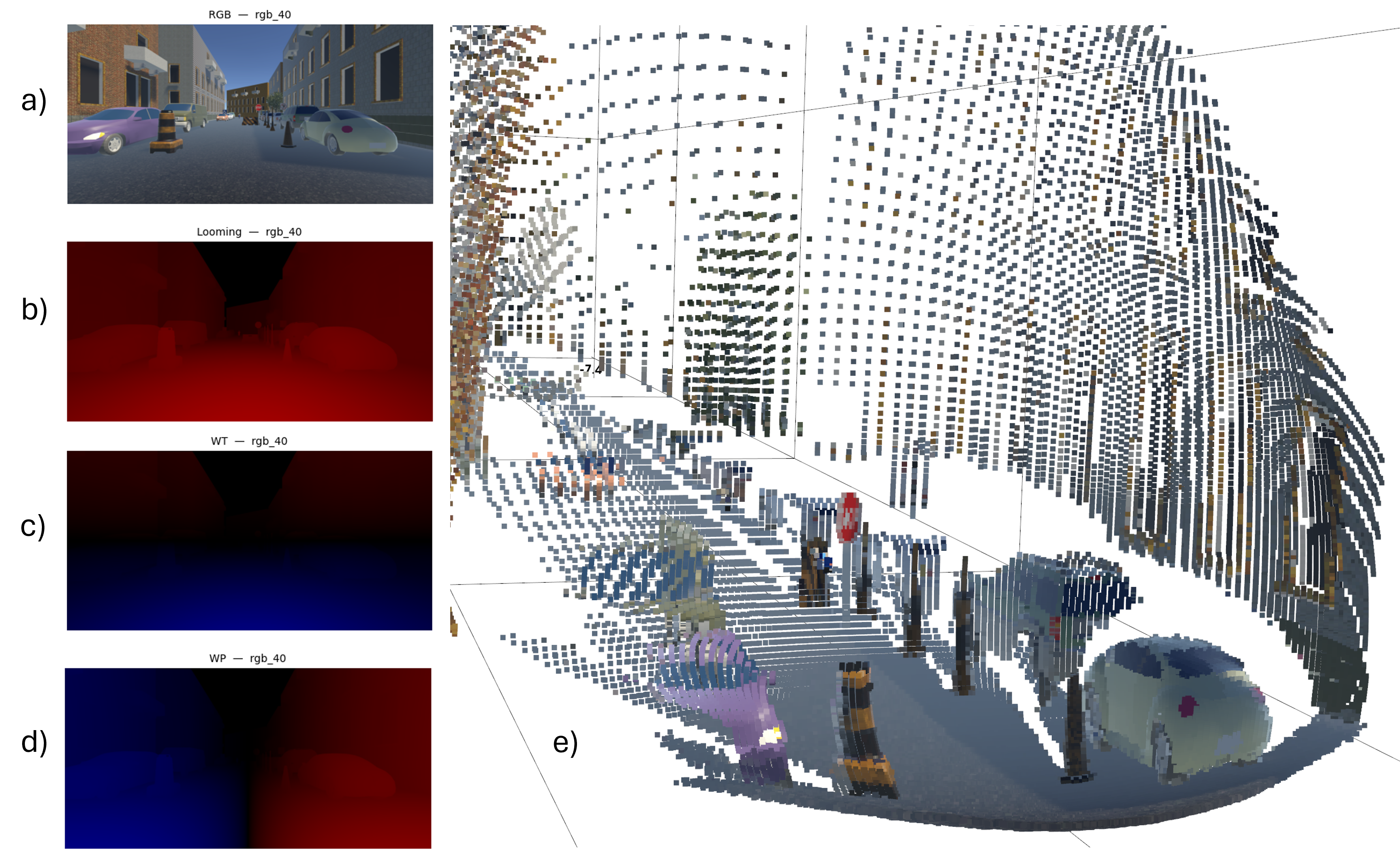}
    \caption{OWL-based 3D reconstruction. (a) RGB image. (b) Looming ($L$). (c)-(d) Rotation components ($\omega$). (e) Scaled 3D point cloud (RoT).}
    \label{fig0052c}
\end{figure}

The second experiment (Figure~\ref{fig0052c}) uses a Unity-based simulation of a camera moving with rectilinear motion along a street scene. The rendered images represent the RGB view observed by the camera. Custom shaders are used to compute, for every pixel, the instantaneous looming value $L$ and the components of the perceived rotation vector $\omega$. For a selected frame, these quantities are exported and processed in Python. Using the measured $L$, $\omega_\theta$, and $\omega_\phi$ values, the quaternion ratio $ToR$ is computed and subsequently converted to $RoT$. The resulting values are then visualized as a scaled 3D point cloud using PyVista. Figure~\ref{fig0052c} illustrates (a) the RGB image, (b) the looming field $L$, (c)--(d) the components of $\omega$, and (e) the reconstructed point cloud in the $RoT$ domain.

\textbf{These results demonstrate that using only the perceived visual motion cues $L$ and $\omega$, the geometric constancy of 3D structures is preserved over time and a scaled point cloud reconstruction can be obtained.}

\section{Conclusions}
This paper introduced \textit{OWL}, a perception-based analytical function that unifies two fundamental visual motion cues, perceived looming ($L$) and perceived rotation ($\omega$), into a single closed-form representation of relative 3D structure. By showing that the quaternion ratio between translation and range can be expressed directly as $L + \boldsymbol{\omega}$, the work establishes a principled link between instantaneous image based perceived visual motion cues and scaled 3D geometry, without explicitly estimating depth, or relying on prior knowledge. The reciprocal \textit{OWL} representation enables scaled 3D reconstruction and preserves geometric constancy of stationary objects over time under relative motion. The framework also provides a direct method for recovering camera heading from the ratio $\omega/L$. The approach extends naturally from 2D complex domain to 3D using quaternions. Simulation results demonstrate shape constancy and point cloud reconstruction from raw visual motion cues alone. Overall, \textit{OWL} offers a minimal, parallel, and perception-grounded alternative to conventional structure-from-motion and learning-based approaches, with implications for robotics, autonomous navigation, and potentially for understanding aspects of natural visual perception.
Currently we are working on expanding the simulation results, adding noise, exploring limitations and using real data.

\bibliographystyle{IEEEtran}
\bibliography{raviv-yepes.bib}

@article{Raviv2000TheVL,
	title={The Visual Looming Navigation Cue: A Unified Approach},
	author={D. Raviv and K. Joarder},
	journal={Comput. Vis. Image Underst.},
	year={2000},
	volume={79},
	pages={331-363}
}

@book{raviv1992quantitative,
	title={A quantitative approach to looming},
	author={Raviv, Daniel},
	year={1992},
	publisher={US Department of Commerce, National Institute of Standards and Technology}
}

@inproceedings{teed2020raft,
	title={Raft: Recurrent all-pairs field transforms for optical flow},
	author={Teed, Zachary and Deng, Jia},
	booktitle={European conference on computer vision},
	pages={402--419},
	year={2020},
	organization={Springer}
}

@book{gibson2014ecological,
	title={The ecological approach to visual perception: classic edition},
	author={Gibson, James J},
	year={2014},
	publisher={Psychology Press}
}

@article{horn1981determining,
	title={Determining optical flow},
	author={Horn, Berthold KP and Schunck, Brian G},
	journal={Artificial intelligence},
	volume={17},
	number={1-3},
	pages={185--203},
	year={1981},
	publisher={Elsevier}
}

@article{lee1976theory,
	title={A theory of visual control of braking based on information about time-to-collision},
	author={Lee, David N},
	journal={Perception},
	volume={5},
	number={4},
	pages={437--459},
	year={1976},
	publisher={SAGE Publications Sage UK: London, England}
}

@book{hartley2003multiple,
	title={Multiple view geometry in computer vision},
	author={Hartley, Richard and Zisserman, Andrew},
	year={2003},
	publisher={Cambridge university press}
}

@book{thomas1994spherical,
  title={Spherical retinal flow for a fixating observer},
  author={Thomas, Inigo and Simoncelli, Eero and Bajcsy, Ruzena},
  year={1994},
  publisher={University of Pennsylvania, School of Engineering and Applied Science~…}
}

@article{longuet1980interpretation,
  title={The interpretation of a moving retinal image},
  author={Longuet-Higgins, Hugh Christopher and Prazdny, Kvetoslav},
  journal={Proceedings of the Royal Society of London. Series B. Biological Sciences},
  volume={208},
  number={1173},
  pages={385--397},
  year={1980},
  publisher={The Royal Society London}
}

@conference{vehits23,
author={Juan Yepes and Daniel Raviv},
title={Visual Looming from Motion Field and Surface Normals},
booktitle={Proceedings of the 9th International Conference on Vehicle Technology and Intelligent Transport Systems - VEHITS},
year={2023},
pages={46-53},
publisher={SciTePress},
organization={INSTICC},
doi={10.5220/0011727400003479},
isbn={978-989-758-652-1},
issn={2184-495X},
}

@misc{raviv2025visionbasedclosedformsolutionmeasuring,
      title={A Vision-Based Closed-Form Solution for Measuring the Rotation Rate of an Object by Tracking One Point}, 
      author={Daniel Raviv and Juan D. Yepes and Eiki M. Martinson},
      year={2025},
      eprint={2507.03237},
      archivePrefix={arXiv},
      primaryClass={cs.CV},
      url={https://arxiv.org/abs/2507.03237}, 
}

@INPROCEEDINGS{323828,
  author={Raviv and Ozery},
  booktitle={1994 Proceedings of IEEE Conference on Computer Vision and Pattern Recognition}, 
  title={A visual-motion fixation invariant}, 
  year={1994},
  volume={},
  number={},
  pages={188-193},
  doi={10.1109/CVPR.1994.323828}}

@article{heeger1992subspace,
  title={Subspace methods for recovering rigid motion I: Algorithm and implementation},
  author={Heeger, David J and Jepson, Allan D},
  journal={International Journal of Computer Vision},
  volume={7},
  number={2},
  pages={95--117},
  year={1992},
  publisher={Springer}
}

@inproceedings{wang2024dust3r,
  title={Dust3r: Geometric 3d vision made easy},
  author={Wang, Shuzhe and Leroy, Vincent and Cabon, Yohann and Chidlovskii, Boris and Revaud, Jerome},
  booktitle={Proceedings of the IEEE/CVF conference on computer vision and pattern recognition},
  pages={20697--20709},
  year={2024}
}

@inproceedings{yang2024depth,
  title={Depth anything: Unleashing the power of large-scale unlabeled data},
  author={Yang, Lihe and Kang, Bingyi and Huang, Zilong and Xu, Xiaogang and Feng, Jiashi and Zhao, Hengshuang},
  booktitle={Proceedings of the IEEE/CVF conference on computer vision and pattern recognition},
  pages={10371--10381},
  year={2024}
}

@article{yang2024depthv2,
  title={Depth anything v2},
  author={Yang, Lihe and Kang, Bingyi and Huang, Zilong and Zhao, Zhen and Xu, Xiaogang and Feng, Jiashi and Zhao, Hengshuang},
  journal={Advances in Neural Information Processing Systems},
  volume={37},
  pages={21875--21911},
  year={2024}
}

@article{kerbl20233d,
  title={3d gaussian splatting for real-time radiance field rendering.},
  author={Kerbl, Bernhard and Kopanas, Georgios and Leimk{\"u}hler, Thomas and Drettakis, George and others},
  journal={ACM Trans. Graph.},
  volume={42},
  number={4},
  pages={139--1},
  year={2023}
}

@article{mildenhall2021nerf,
  title={Nerf: Representing scenes as neural radiance fields for view synthesis},
  author={Mildenhall, Ben and Srinivasan, Pratul P and Tancik, Matthew and Barron, Jonathan T and Ramamoorthi, Ravi and Ng, Ren},
  journal={Communications of the ACM},
  volume={65},
  number={1},
  pages={99--106},
  year={2021},
  publisher={ACM New York, NY, USA}
}

@book{churchill2013complex,
  author    = {R. V. Churchill and J. W. Brown},
  title     = {Complex Variables and Applications},
  edition   = {9th},
  publisher = {McGraw-Hill},
  year      = {2013}
}

\section*{Appendix A: The relation between $\omega$, optical flow, and camera rotation \(\Omega\) from the Motion Field}

The motion field of a 3D point $\mathbf{r}$ in camera coordinates, due to camera
instantaneous translation $\mathbf{t}$ and rotation $\boldsymbol{\Omega}$, is given
by \cite{longuet1980interpretation}:
\begin{figure}[htpb]
    \centering
    \includegraphics[width=0.3\textwidth]{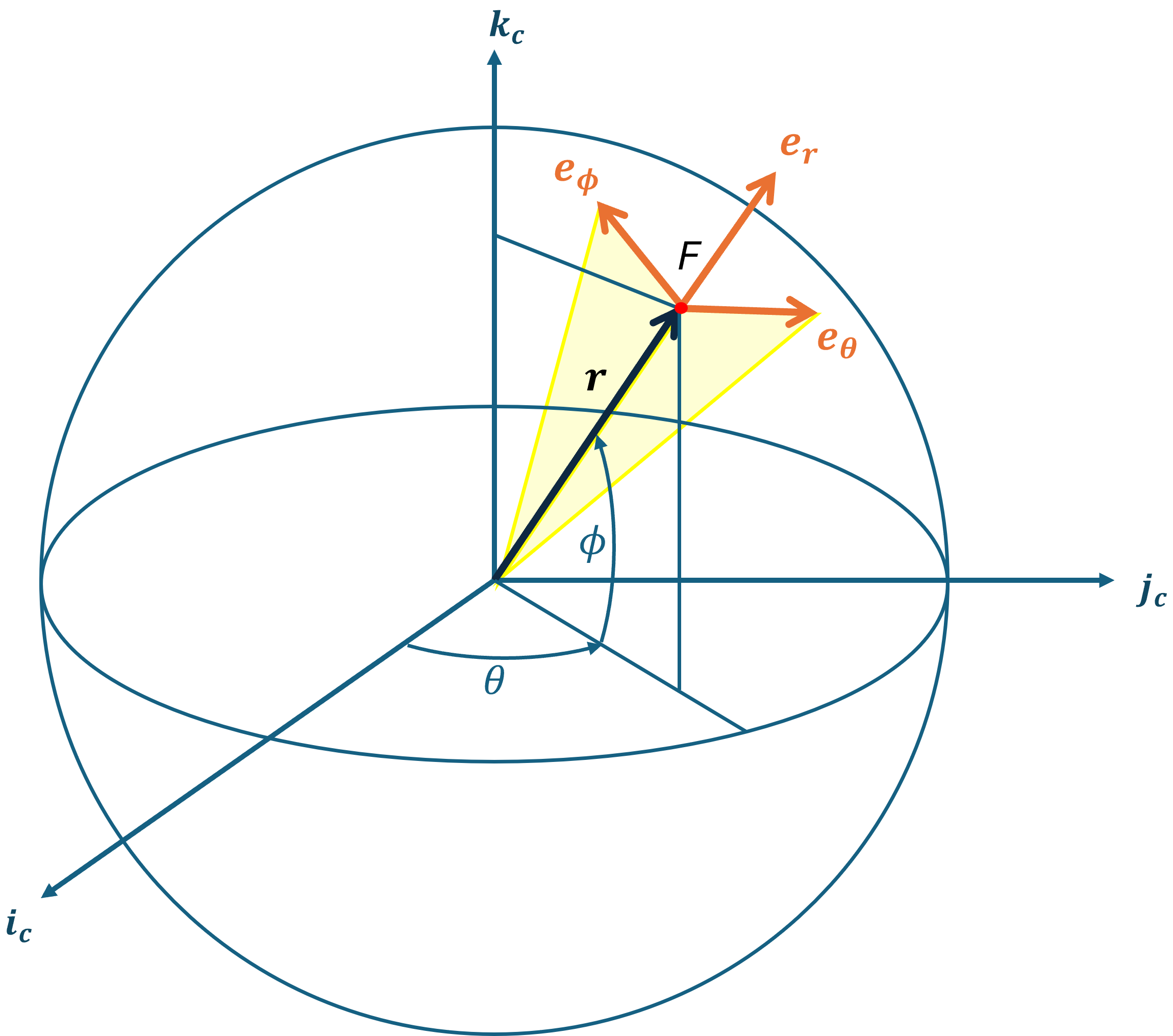}
    \caption{Coordinate system and directional unit vectors}
    \label{fig0052}
\end{figure}

\begin{align}
    \mathbf{V}_F = -\mathbf{t} - \boldsymbol{\Omega} \times \mathbf{r}
    \label{eq:VF_rigid}
\end{align}

Where the same velocity can be written in spherical coordinates $(r,\theta,\phi)$ as (see Figure \ref{fig0052})

\begin{align}
    \mathbf{V}_F = \dot{r}\,\mathbf{e}_r + r\dot{\theta}\cos\phi\,\mathbf{e}_{\theta}
                  + r\dot{\phi}\,\mathbf{e}_{\phi}.
    \label{eq:VF_spherical}
\end{align}

Equating \eqref{eq:VF_rigid} and \eqref{eq:VF_spherical}, dividing both sides by $|\mathbf{r}|=r$ and using
$(-\boldsymbol{\Omega}\times\mathbf{r})/r = \mathbf{e}_r\times\boldsymbol{\Omega}$
yields:

\begin{align}
    \frac{-\mathbf{t}}{r} + \mathbf{e}_r\times\boldsymbol{\Omega}
    = \frac{\dot{r}}{r}\,\mathbf{e}_r
      + \dot{\theta}\cos\phi\,\mathbf{e}_{\theta}
      + \dot{\phi}\,\mathbf{e}_{\phi}.
    \label{eq:divided}
\end{align}

Decomposing $\mathbf{t}$ and $\boldsymbol{\Omega}$ along the local spherical basis,

\begin{align}
    \mathbf{t} &= t_r\mathbf{e}_r + t_\theta\mathbf{e}_\theta + t_\phi\mathbf{e}_\phi,
    \qquad
    \boldsymbol{\Omega} = \Omega_r\mathbf{e}_r
                         + \Omega_\theta\mathbf{e}_\theta
                         + \Omega_\phi\mathbf{e}_\phi,
\end{align}

and noting that $\mathbf{e}_r\times\boldsymbol{\Omega}
= \Omega_\theta\,\mathbf{e}_\phi - \Omega_\phi\,\mathbf{e}_\theta$,
equation \eqref{eq:divided} separates by component:

\begin{align}
    \mathbf{e}_r:&\quad
        \frac{\dot{r}}{r} = -\frac{t_r}{r}
        \;\;\Longrightarrow\;\;
        L \equiv -\frac{\dot{r}}{r} = \frac{\mathbf{t}\cdot\mathbf{e}_r}{|\mathbf{r}|},
        \label{eq:L}\\[4pt]
    \mathbf{e}_\theta:&\quad
        \dot{\theta}\cos\phi = -\frac{t_\theta}{r} - \Omega_\phi
        = -\frac{\mathbf{t}\cdot\mathbf{e}_\theta}{|\mathbf{r}|} - \Omega_\phi,
        \label{eq:thetadot}\\[4pt]
    \mathbf{e}_\phi:&\quad
        \dot{\phi} = -\frac{t_\phi}{r} + \Omega_\theta
        = -\frac{\mathbf{t}\cdot\mathbf{e}_\phi}{|\mathbf{r}|} + \Omega_\theta.
        \label{eq:phidot}
\end{align}

The translational contribution to the angular flow, which we refer to the perceived $\boldsymbol{\omega}$ in vector form has components

\begin{align}
    \omega_\phi \equiv \frac{\mathbf{t}\cdot\mathbf{e}_\theta}{|\mathbf{r}|}, \qquad
    \omega_\theta \equiv \frac{-\mathbf{t}\cdot\mathbf{e}_\phi}{|\mathbf{r}|},
    \label{eq:omega_components}
\end{align}

equations \eqref{eq:thetadot} and \eqref{eq:phidot} can be reduced to:

\begin{align}
    \dot{\theta}\cos\phi &= -(\omega_\phi + \Omega_\phi), \label{eq:thetadot_final}\\
    \dot{\phi}           &=   \omega_\theta + \Omega_\theta. \label{eq:phidot_final}
\end{align}

\section*{Appendix B: Recovering $\hat{\mathbf{t}}$ from $L$ and $\boldsymbol{\omega}$ via Quaternions}

Let's represent the range and translation vectors as pure quaternions, i.e.,
$R=(0,\mathbf{r})$ and $T=(0,\mathbf{t})$.
The inverse of a pure quaternion is $R^{-1} = (0,\mathbf{r}/|\mathbf{r}|^2)$, and so the product of $T$ and $R^{-1}$ becomes:

\begin{align}
    ToR = T \otimes R^{-1}
        = (0,\mathbf{t})\otimes\!\left(0,\tfrac{-\mathbf{r}}{|\mathbf{r}|^2}\right).
    \label{eq:ToR_def}
\end{align}

Applying the quaternion product rule
$(0,\mathbf{a})\otimes(0,\mathbf{b})=(-\mathbf{a}\cdot\mathbf{b},\;\mathbf{a}\times\mathbf{b})$:

\begin{align}
    ToR = \left(\frac{\mathbf{t}\cdot\mathbf{e}_r}{|\mathbf{r}|},\;
            \frac{\mathbf{t}\times(-\mathbf{e}_r)}{|\mathbf{r}|}\right)
    = L + \boldsymbol{\omega}
    \label{eq:ToR_Lw}
\end{align}

where the scalar part $L=\mathbf{t}\cdot\mathbf{e}_r/|\mathbf{r}|$ is the looming
term \eqref{eq:L} and the vector part
$\boldsymbol{\omega}=\mathbf{t}\times(-\mathbf{e}_r)/|\mathbf{r}|$
is the derotated flow \eqref{eq:omega_components}.

\medskip
The unit translation direction follows by acting $ToR$ on $\mathbf{e}_r$ and
normalising. Since $\boldsymbol{\omega}\perp\mathbf{e}_r$, one has
$\boldsymbol{\omega} = -\mathbf{t}\times\mathbf{e}_r/|\mathbf{r}|
                     = \boldsymbol{\omega}\times\mathbf{e}_r\times\mathbf{e}_r$,
and therefore $\mathbf{t}/|\mathbf{r}| = L\,\mathbf{e}_r + \boldsymbol{\omega}\times\mathbf{e}_r$.
Normalising:

\begin{align}
    \hat{\mathbf{t}} =
        \frac{L\,\mathbf{e}_r + \boldsymbol{\omega}\times\mathbf{e}_r}
             {\sqrt{L^2 + |\boldsymbol{\omega}|^2}}
    \label{eq:that}
\end{align}

Thus $\hat{\mathbf{t}}$ is fully encoded in $ToR$, with no dependence on
$\boldsymbol{\Omega}$. The angle $\alpha$ between $\hat{\mathbf{t}}$ and the line
of sight satisfies $\tan\alpha = |\boldsymbol{\omega}|/L = \|\operatorname{Im}(ToR)\|/\operatorname{Re}(ToR)$.

\end{document}